\documentclass[lettersize,journal]{IEEEtran}
\usepackage{amsmath,amsfonts}
\usepackage{algorithm}
\usepackage{algorithmic}
\usepackage{array}
\usepackage[caption=false,font=normalsize,labelfont=sf,textfont=sf]{subfig}
\usepackage{textcomp}
\usepackage{stfloats}
\usepackage{url}
\usepackage{verbatim}
\usepackage{graphicx}
\usepackage{cite}
\usepackage{booktabs}
\usepackage{multirow}   
\usepackage{amsmath}
\usepackage{amssymb}
\begin{document}

\title{Polarized Direct Cross-Attention Message Passing in GNNs for Machinery Fault Diagnosis}

\author{Zongyu Shi, Laibin Zhang, Maoyin Chen
\thanks{Zongyu Shi is with the Department of Automation, College of Artificial Intelligence, China University of Petroleum (Beijing), Beijing 102299, China}
\thanks{Laibin Zhang and Maoyin Chen are with the Department of Safety Engineering, College of Safety and Ocean Engineering, China University of Petroleum (Beijing), Beijing 102299, China}
}

\maketitle

\begin{abstract} The reliability of safety-critical industrial systems hinges on accurate and robust fault diagnosis in rotating machinery. Conventional graph neural networks (GNNs) for machinery fault diagnosis face limitations in modeling complex dynamic interactions due to their reliance on predefined static graph structures and homogeneous aggregation schemes. To overcome these challenges, this paper introduces polarized direct cross-attention (PolaDCA), a novel relational learning framework that enables adaptive message passing through data-driven graph construction. Our approach builds upon a direct cross-attention (DCA) mechanism that dynamically infers attention weights from three semantically distinct node features (such as individual characteristics, neighborhood consensus, and neighborhood diversity) without requiring fixed adjacency matrices.
Theoretical analysis establishes PolaDCA's superior noise robustness over conventional GNNs.
Extensive experiments on industrial datasets (i.e., XJTUSuprgear, CWRUBearing and Three-Phase Flow Facility datasets) demonstrate state-of-the-art diagnostic accuracy and enhanced generalization under varying noise conditions, outperforming seven competitive baseline methods. The proposed framework provides an effective solution for safety-critical industrial applications.
\end{abstract}


\section{Introduction}
\label{sec:introduction}
\IEEEPARstart{M}{odern} industrial systems rely heavily on critical rotating equipment like bearings and compressors. These components frequently operate in demanding, high-load environments and are susceptible to malfunctions. Consequently, machinery fault diagnosis has emerged as a vital technology for risk reduction in safety-critical applications. Unreliable fault diagnosis can lead to missed alarms or false positives, directly impacting system availability, maintenance costs, and operational safety. To achieve higher accuracy and generalization in fault diagnosis, research has increasingly turned to advanced deep learning models. Among these, graph neural networks (GNNs) \cite{GNN} hold unique importance because they explicitly address a key aspect of machinery systems: interconnectedness. Unlike methods that process data points in isolation \cite{Review}, GNNs models the entire sensor network as a graph, learning how a fault in one component propagates and affects related nodes. This makes GNNs exceptionally suited for exact fault diagnosis.

In recent years, GNNs have emerged as a dominant paradigm by leveraging structural dependencies. Early architectures, such as graph convolutional networks (GCN) \cite{GCN} and graph sample and aggregate (GraphSAGE) \cite{GraphSAGE}, perform neighborhood aggregation through spectral approximations and inductive sampling frameworks, respectively. To enhance representational capacity, graph attention networks (GAT) introduce localized attention mechanisms that assign non-uniform importance to neighbors \cite{GAT}, while graph transformer (GTF) expands the scope to global dependencies by self-attention \cite{GTF}, effectively modeling long-range interactions across the entire graph. 
Graph contrastive learning (GCL) leverages contrastive learning for robust node representations \cite{GCL_main,GCL_2}. Multireceptive field graph convolutional
networks (MRF-GCN) focus on multi-scale feature extraction \cite{MRFGNN}, while interaction-aware GNNs (IAGNN) \cite{IAGNN} considers the awareness of heterogeneous node-edge interactions. These works provide a more granular understanding of network topologies.

Now GNNs have significantly promoted the development of fault diagnosis by modeling topological dependencies among multi-sensors signals. Unlike traditional signal processing methods \cite{Review}, GNNs effectively capture spatial correlations between disparate sensing components and the temporal dynamics of system states. The state-of-the-art has recently been expanded by works such as hypergraph-based multisensor fusion to capture high-order correlations in data \cite{Yan2024}. To tackle the distribution discrepancy across different devices, a dual-contrastive multi-view GAT is designed for cross-machine diagnosis under domain and label shifts \cite{Zhu2025}. To address the integration of domain expertise, a graph convolutional neural network leverages knowledge graphs to embed structured engineering semantics into the learning process \cite{Zhao2024_KG}.

Current research also considers improving the interpretability and robustness of diagnosis by integrating prior physical knowledge into graph construction or employing advanced reasoning modules. Specifically, causal disentangled GNN (CDGNN) leverages causal theory to disentangle causal substructures from bias substructures \cite{CDGNN}, effectively mitigating the misleading effects of spurious correlations. Fuzzy inference-guided GNN (FIGNN) use fuzzy inference to extract causal features \cite{FIGNN}, bridging the gap between symbolic reasoning and neural learning. For dynamic modeling, multiattention spatiotemporal fusion GNN (MA-STGNN) integrates spatiotemporal fusion with a multi-head self-attention to capture the synchronous and asynchronous evolution of fault signatures \cite{MA-STGNN}. These approaches enable GNNs to distinguish between transient disturbances and genuine structural failures, thereby providing reliable decision support for predictive maintenance in industrial systems.

Although existing GNN-based methods have improved diagnostic performance, the inherent message-passing mechanisms have fundamental limitations. First, they heavily rely on predefined static graph structures.  
Second, they are mostly local neighborhood aggregation operations, which incur information loss by discarding higher-order statistical features and struggle to efficiently model long-range dependencies among components. 
To address these two limitations, we abandon the fixed-graph-dependent message passing and instead adopt a data-driven DCA to dynamically establish a fully-connected attention graph based on content similarity. Most of all, we introduce the polarized DCA (PolaDCA), which models the positive/negative polarity (enhancing/suppressing) of interactions to finely characterize the complex physical effects in fault propagation. The main merits are listed as follows:

(1) Feature-driven graph learning. Unlike conventional GNNs that depend on fixed adjacency matrices or predefined topologies, we introduce a DCA-driven message-passing where the graph is constructed entirely from node features. The DCA is mainly based on semantic similarity between different feature representations (individual node characteristics, neighborhood consensus, and neighborhood diversity). This enables GNNs to learn interaction patterns, resulting in a more flexible approach for relational reasoning. 

(2) Polarity-aware interaction. We extend the concept of attention beyond weighting and introduce explicit modeling of interaction polarity. By decomposing attention into positive (enhancing) and negative (suppressing) components, the PolaDCA allows the model to distinguish not only how strongly nodes interact, but also whether their influence is synergistic or antagonistic—providing a more accurate representation of fault propagation mechanisms in machinery systems.

The remainder of this paper is organized as follows: Section II formulates the problem and preliminary. Section III presents the DCA mechanism for message passing. Section IV extends DCA to PolaDCA by introducing polarity-aware interactions. Section V provides a theoretical analysis on noise robustness. Section VI describes DCA-/PolaDCA-based diagnosis framework. Section VII presents experimental results on three industrial datasets. Section VIII concludes with a discussion of limitations and future directions. 

\section{Problem Formulation and Preliminaries}

\subsection{Problem Formulation}

This research mainly considers the problem of machinery fault diagnosis. Given a vibration signal sequence $\boldsymbol{X} \in \boldsymbol{R}^{m\times L}$ where $m$ is the number of sensors and $L$ is the length, our goal is to learn a mapping function $\boldsymbol{\Phi}: \boldsymbol{X} \rightarrow y$ that classifies it into one of $K$ predefined condition categories $y \in \{1, 2, \dots, K\}$, thereby enabling high-precision condition monitoring and early fault warning.

To address the limitations of message passing schemes in current GNNs, we want to transform the fault diagnosis from a ``local information aggregation" constrained by fixed structures and connections into a ``relation reasoning" capable of autonomously leveraging deep semantic relationships among node signals. This will be realized by DCA, which dynamically infers interaction strengths across node features through attention weights, eliminating the need for any predefined graph adjacency. More importantly, we extend this mechanism to PolaDCA, which additionally learns the polarization (positive or negative) of each interaction.

\subsection{Preliminaries}
\subsubsection{Graph convolutional networks (GCNs)\cite{GCN}}

GCNS employ a spectral-based convolution operation approximated through first-order neighborhood aggregation. Given a graph $\mathcal{G} = (\mathcal{V}, \mathcal{E})$, where $N$ is the number of nodes and $\mathcal{N}(i)$ represents the neighborhood of node $v_i$. The layer-wise propagation rule is defined as
\begin{equation}
\label{layer-propa}
\boldsymbol{H}^{(l+1)} = \sigma\left(\boldsymbol{D}^{-\frac{1}{2}}\tilde{\boldsymbol{A}}\boldsymbol{D}^{-\frac{1}{2}}\boldsymbol{H}^{(l)}\boldsymbol{W}^{(l)}\right)
\end{equation}
where $\tilde{\boldsymbol{A}}=\boldsymbol{A}+\boldsymbol{I}_N$ represents the adjacency matrix with added self-loops, $\boldsymbol{D}$ is the corresponding degree matrix with $D_{ii}=\sum_j\tilde{A}_{ij}$, $\boldsymbol{W}^{(l)}\in\mathbb{R}^{D^{(l)}\times D^{(l+1)}}$ denotes the trainable weight matrix at layer $l$, and $\sigma(\cdot)$ is an activation function. Eq.~(\ref{layer-propa}) is equivalently expressed at the node level as
\begin{equation}
\label{GCNs}
\boldsymbol{h}_i^{(l+1)} = \sigma\left(\sum_{j\in\mathcal{N}(i)\cup{i}}\frac{1}{\sqrt{d_id_j}}\boldsymbol{W}^{(l)}\boldsymbol{h}_j^{(l)}\right)
\end{equation}
where $d_i$ indicates the degree of node $i$ including self-loops. However, the limitation of this approach lies in its fixed aggregation weights determined solely by node degrees. 

\subsubsection{Graph attention networks (GATs)\cite{GAT}} GATs address the above limitation by learning attention weights through a self-attention mechanism. The attention coefficients between nodes $v_i$ and $v_j$ are computed as
\begin{equation}
e_{ij} = \text{LeakyReLU}\left(\boldsymbol{a}^T[\boldsymbol{W} \boldsymbol{h}_i | \boldsymbol{W} \boldsymbol{h}_j]\right)
\end{equation}
where $\boldsymbol{a}\in\mathbb{R}^{2D'}$ is a learnable attention vector, $\boldsymbol{W}\in\mathbb{R}^{D'\times D}$ is a weight matrix, and `$|$' denotes concatenation. These coefficients are normalized using the softmax function:
\begin{equation}
\alpha_{ij} = \frac{\exp(e_{ij})}{\sum_{k\in\mathcal{N}(i)}\exp(e_{ik})}
\end{equation}
The node representation is then updated as
\begin{equation}
\boldsymbol{h}_i' = \sigma\left(\sum_{j\in\mathcal{N}(i)}\alpha_{ij}\boldsymbol{W} \boldsymbol{h}_j\right)
\end{equation}
While providing adaptive weighting of neighbors, it remains constrained to capturing only pairwise similarity relationships, ignoring the diverse interaction patterns (such as synergistic reinforcement or compensatory effects). 

\subsubsection{Standard cross-attention (SCA)}
Following the cross-modal fusion paradigm \cite{Transformer_SCA}, SCA first aligns features from different modalities into a common latent semantic space. Given a Query feature $\boldsymbol{X}_E \in \mathbb{R}^{n_X \times d_X}$ and a context feature $\boldsymbol{Y}_D \in \mathbb{R}^{n_Y \times d_Y}$, modality-specific projection heads $P_X: \mathbb{R}^{d_X} \to \mathbb{R}^{d_Z}$ and $P_Y: \mathbb{R}^{d_Y} \to \mathbb{R}^{d_Z}$ map them into the shared space $\mathcal{Z}$, i.e.,
$\tilde{\boldsymbol{X}}_E = P_X(\boldsymbol{X}_E) \in \mathbb{R}^{n_X \times d_Z}$, $\tilde{\boldsymbol{Y}}_D = P_Y(\boldsymbol{Y}_D) \in \mathbb{R}^{n_Y \times d_Z}$. Then, within $\mathcal{Z}$, learnable matrices transform the aligned features into Query, Key, and Value representations:
\begin{align}
    \boldsymbol{Q} &= \tilde{\boldsymbol{X}}_E \boldsymbol{W}^Q \in \mathbb{R}^{n_X \times d_q}, 
    & \boldsymbol{W}^Q &\in \mathbb{R}^{d_Z \times d_q}, \label{eq:Q} \\
    \boldsymbol{K} &= \tilde{\boldsymbol{Y}}_D \boldsymbol{W}^K \in \mathbb{R}^{n_Y \times d_k}, 
    & \boldsymbol{W}^K &\in \mathbb{R}^{d_Z \times d_k}, \label{eq:K} \\
    \boldsymbol{V} &= \tilde{\boldsymbol{Y}}_D \boldsymbol{W}^V \in \mathbb{R}^{n_Y \times d_v}, 
    & \boldsymbol{W}^V &\in \mathbb{R}^{d_Z \times d_v}. \label{eq:V}
\end{align}
Finally, the scaled dot-product attention is computed as
\begin{equation}
    \text{Attention}(\boldsymbol{Q},\boldsymbol{K},\boldsymbol{V}) = 
    \text{softmax}\!\left(\frac{\boldsymbol{Q} \boldsymbol{K}^T}{\sqrt{d_k}}\right) \boldsymbol{V}
    \label{eq:sca_attention}
\end{equation}
where $d_k$ is the dimension of the Key vectors.

\section{DCA-based Message Passing}
\label{DCASection}

Here, we develop DCA-based message passing by dynamically learning correlations from neighboring node features. The DCA enables a nuanced and adaptive representation of local graph structures.
The overall architecture with DCA-based message passing is depicted in Fig. 1, which illustrates the DCA-based message passing and dynamic gating mechanism. 

\begin{figure*}[t]       
  \centering
  \includegraphics[width=5in]{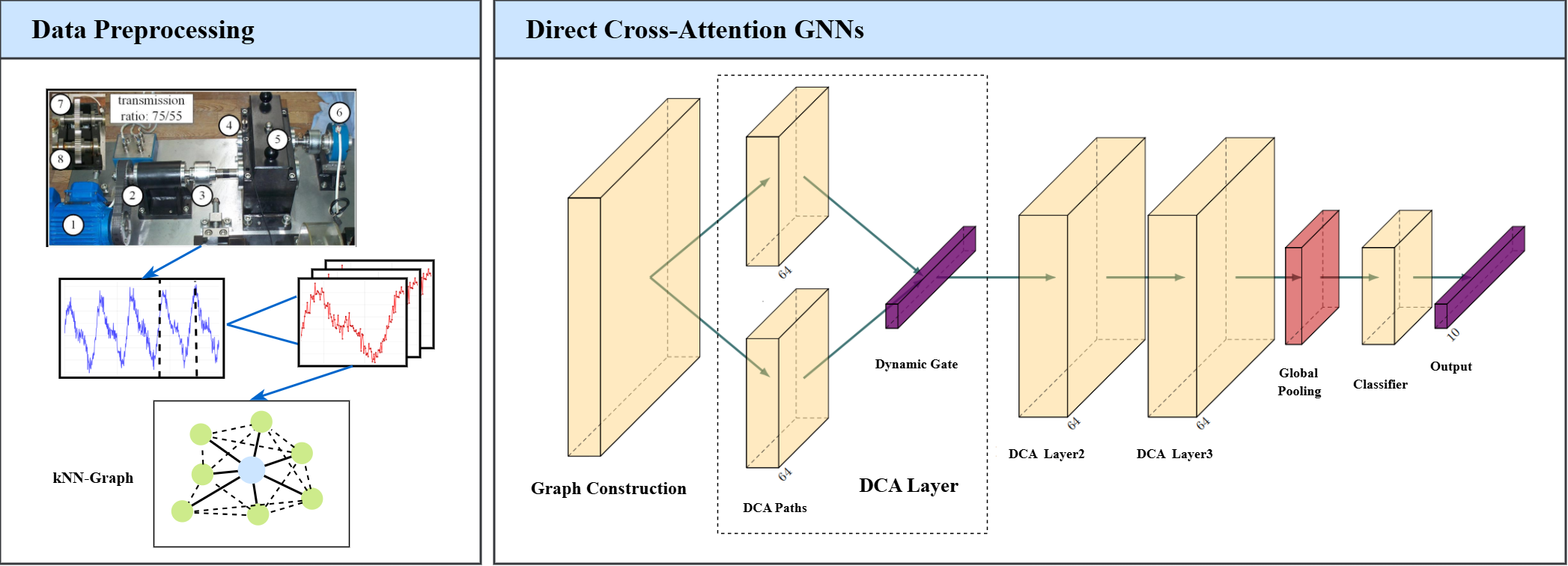}
  \caption{Framework of DCA-GNN}
  \label{framework}
\end{figure*}

\subsection{Direct Cross-Attention (DCA)}

We first introduce the definition of the DCA operator.

\textbf{Definition 1 (DCA)}: Given a heterogeneous feature set \( \mathcal{H} = \{\boldsymbol{f}_i, \boldsymbol{f}_j, \boldsymbol{f}_k\} \), where $\boldsymbol{f}_i \in \mathbb{R}^{1 \times d_i}$, $\boldsymbol{f}_j \in \mathbb{R}^{1 \times d_j}$ and $\boldsymbol{f}_k\in \mathbb{R}^{1 \times d_k}$. Unlike SCA, DCA does not employ a shared homogenization projection to a common latent space. Instead, it enables interactions across original feature spaces through dedicated, path-specific projection matrices:
\begin{equation}
\begin{split}
    &\text{DCA}(\boldsymbol{f}_i, \boldsymbol{f}_j, \boldsymbol{f}_k; \boldsymbol{W}_{f_i}^Q, \boldsymbol{W}_{f_j}^K, \boldsymbol{W}_{f_k}^V )  
    \\ &= \text{softmax} \left( \frac{( \boldsymbol{f}_i\boldsymbol{W}_{f_i}^Q)( \boldsymbol{f}_j\boldsymbol{W}_{f_j}^K)^T}{\sqrt{d_k}} \right) ( \boldsymbol{f}_k\boldsymbol{W}_{f_k}^V)
    \label{DCA}
\end{split}
\end{equation}
where \( \boldsymbol{W}_{f_i}^Q \in \mathbb{R}^{d_i \times d_q}, \boldsymbol{W}_{f_j}^K \in \mathbb{R}^{d_j \times d_k}, \boldsymbol{W}_{f_k}^V \in \mathbb{R}^{d_k \times d_v} \) are learnable projection matrices, depending on the configurations of Query, Key, and Value in the attention.

In contrast to SCA, the DCA operator allows features of different spaces to interact directly without homogenization projections \( P_m \). It should be noted that DCA still employs learnable projection matrices \( \boldsymbol{W}_{f_i}^Q, \boldsymbol{W}_{f_j}^K, \boldsymbol{W}_{f_k}^V \) for each interaction path. These can be viewed as dynamic, path-dependent alignment operations, offering greater flexibility than a fixed shared projection. In addition, the SCA uses two features: one for the Query and the other for both Key and Value. However, DCA explicitly uses three separate features—one for Query, one for Key, and one for Value—enabling richer, decoupled multimodal interactions in a single operation.

\subsection{DCA for Message Passing in GNNs}

\begin{figure*}[t]       
  \centering
  \includegraphics[width=4.5in]{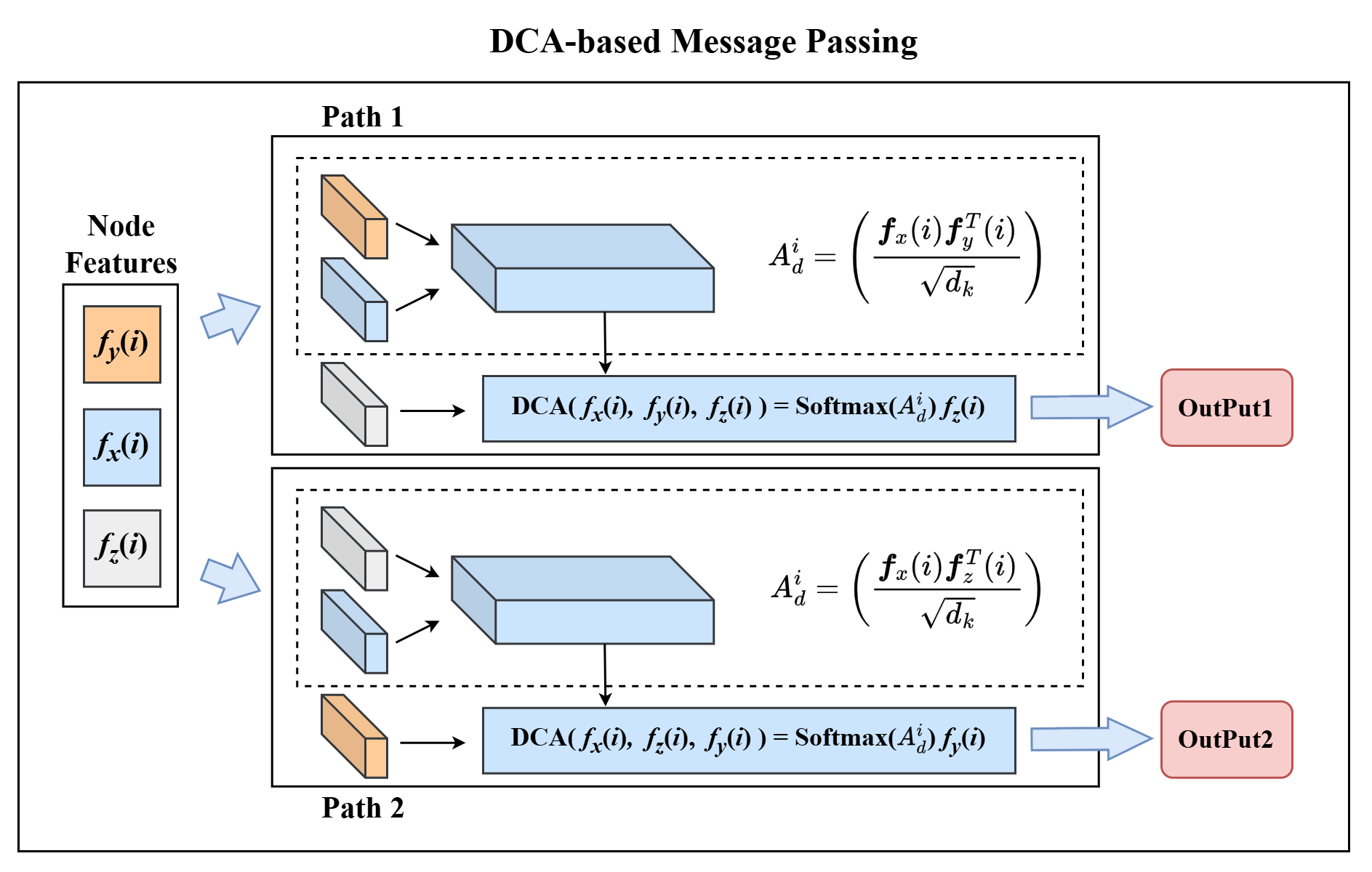}
  \caption{Framework of DCA-based message passing}
  \label{DCA_Main}
\end{figure*}

Given a graph $\mathcal{G} = (\mathcal{V}, \mathcal{E})$ with node $v_i$ and the corresponding node feature $\boldsymbol{x}_i\in \mathbb{R}^{1 \times D} $. Let the node feature matrix $\boldsymbol{N} = [\boldsymbol{x}_1,\boldsymbol{x}_2,\cdots,\boldsymbol{x}_n] \in \mathbb{R}^{n \times D}$, where $n$ is the number of nodes, $D$ is the feature dimension, and $\mathcal{N}(i)$ represent the neighborhood of node $v_i$. 

For node $v_i \in \mathcal{V}$, we define the node feature:
\begin{equation}
   \boldsymbol{f}_x(i) = \boldsymbol{x}_i\boldsymbol{W}_x^i 
    \label{X}
\end{equation} 
where $\boldsymbol{W}_x^i \in \mathbb{R}^{D \times M}$ is a learnable projection matrix, strongly depending on the feature of node $v_i$. Next, two types of neighborhood representations for node $v_i$ can be computed. 
One is the neighborhood average feature $\boldsymbol{f}_y(i)$, which represents the consensus information:
\begin{equation}
\label{consensus}
\boldsymbol{f}_y(i)=\frac{1}{|\mathcal{N}(i)|}\sum_{j\in\mathcal{N}(i)}\boldsymbol{x}_j\boldsymbol{W}_y^i
\end{equation}
where $\boldsymbol{W}_y^i \in \mathbb{R}^{D \times M}$ is a learnable projection matrix.
The other is the neighborhood diversity feature $\boldsymbol{f}_z(i)$, capturing the variability among the neighborhood:
\begin{equation}
\label{diversity}
\boldsymbol{f}_z(i)=\sqrt{\frac{\sum_{j\in\mathcal{N}(i)}\left(\boldsymbol{x}_j\boldsymbol{W}_z^i-\boldsymbol{f}_y(i)\right) \odot \left(\boldsymbol{x}_j\boldsymbol{W}_z^i-\boldsymbol{f}_y(i)\right)}{|\mathcal{N}(i)|-1}}
\end{equation}
where `$\odot$' is the Hadamard product, $\boldsymbol{W}_z^i \in \mathbb{R}^{D \times M}$ is a learnable projection matrix, and the square and square root operations are applied element-wise.

Note that the extracted features described above do not follow the standard SCA procedure of first undergoing projection alignment to enter a common semantic space. Although projections are applied to all three features in DCA, this operation merely serves the necessary purpose of dimensionality matching for interaction. In other words, the features from the node and its neighbors do not interact within a unified semantic space, but rather interact directly. 

In contrast to SCA, the above features do not employ a shared homogenization projection to a common latent space. From Eqs.~(\ref{X})-(\ref{diversity}), the projections $\boldsymbol{W}_x^i,\boldsymbol{W}_y^i,\boldsymbol{W}_z^i$ have been incorporated into the features $\boldsymbol{f}_x(i),\boldsymbol{f}_y(i),\boldsymbol{f}_z(i)$, respectively. Hence, the definition of DCA (Eq.~(\ref{DCA})) should be modified for applicability in GNNs. 

\textbf{Definition 2 (DCA for GNNs)}: For node $v_i$ in GNN, given a set of features \( \mathcal{H}_i = \{\boldsymbol{f}_x(i), \boldsymbol{f}_y(i), \boldsymbol{f}_z(i)\} \) with $ \boldsymbol{W}_x^i,\boldsymbol{W}_y^i,\boldsymbol{W}_z^i$ being projection matrices.
The DCA for GNNs is defined by 

\begin{equation}
    \text{DCA}(\boldsymbol{f}_x(i), \boldsymbol{f}_y(i), \boldsymbol{f}_z(i))  
    = \text{softmax} \left( \frac{ \boldsymbol{f}_x(i)\boldsymbol{f}_y^T(i)}{\sqrt{d_k}} \right) \boldsymbol{f}_z(i)
    \label{MDCA}
\end{equation}
where \( \boldsymbol{W}_x^i, \boldsymbol{W}_y^i, \boldsymbol{W}_z^i \) are learnable projection matrices, depending on the node feature, the average feature and the diversity feature of node $v_i$ in GNNs.

In Eq.~(\ref{MDCA}), 
\(\boldsymbol{W}_x^i\), \(\boldsymbol{W}_y^i\), \(\boldsymbol{W}_z^i\) are not fixed but are adaptive, meaning their parameters are learned to generate the most task-relevant representations. 
This design enables the model to learn context-sensitive projections that perform the flow and fusion of information between various feature sources. Thus, DCA (Eq.~(\ref{MDCA})) is chosen as the message passing strategy in GNNs.

\subsection{Dynamic Gating for Relational Reasoning}

Together with DCA for GNNs (Eq.(\ref{MDCA})), featurs $\boldsymbol{f}_x(i)$, $\boldsymbol{f}_y(i)$, $\boldsymbol{f}_z(i)$ enable a comprehensive analysis of node-neighborhood relationships through the following two reasoning paths (as illustrated in Fig.~\ref{DCA_Main}).

(1) Path 1: Consensus-oriented reasoning. 

This investigates how individual nodes relate to the collective behavior of their neighborhoods while considering neighborhood diversity as contextual information:
\begin{equation}
\label{path1}
   \mathrm{Path}_1 = \mathrm{DCA}(\boldsymbol{f}_x(i), \boldsymbol{f}_y(i), \boldsymbol{f}_z(i)) 
\end{equation}

This reasoning is particularly valuable for identifying nodes that deviate from local trends, potentially indicating fault initiation points.

(2) Path 2: Diversity-oriented reasoning. 

This examines how nodes interact with the variability in their neighborhoods, using the neighborhood consensus as a reference frame:
\begin{equation}
\label{path2} 
  \mathrm{Path}_2 = \mathrm{DCA}(\boldsymbol{f}_x(i), \boldsymbol{f}_z(i), \boldsymbol{f}_y(i))  
\end{equation}

It helps identify nodes that maintain consistent behavior despite surrounding variations, indicating robust components or stable reference points.

These two reasoning channels provide complementary perspectives on node-neighborhood relationships. To adaptively integrate these insights based on the specific relational context, we introduce a dynamic gating mechanism:
\begin{equation}
  \label{deqn_exla}
  \begin{gathered}
    g=\sigma(\boldsymbol{W}_g[\mathrm{Path}_1\parallel\mathrm{Path}_2]+\boldsymbol{b}_g)\\[2pt]
    \mathrm{Fused}=g\odot\mathrm{Path}_1+(1-g)\odot\mathrm{Path}_2 
  \end{gathered}
\end{equation}
where $\sigma$ is the sigmoid function, `$\parallel$' denotes concatenation, and $\boldsymbol{W}_{g}\in \mathbb{R}^{M\times 2M}$, $\boldsymbol{b}_{g}\in \mathbb{R}^{M}$ are learnable parameters. This gating mechanism enables the model to (i)  prioritize consensus reasoning when fault patterns manifest as coordinated deviations across multiple sensors, and (ii) 
prioritize diversity reasoning when faults create localized anomalies amidst normal operations.

To further refine the learned relationships and capture specialized interaction patterns, we introduce a multi-expert architecture with feature enhancement connections:
\begin{equation}
  \label{deqn_exla}
  \begin{gathered}
    \alpha=\sigma\left(\frac{1}{E}\sum_{e=1}^{E}r_{e}\right)\\[2pt]
    \mathrm{Output}=\boldsymbol{f}_{x}(i)+\alpha\odot\left(\sum_{e=1}^{E}w_{e}\cdot\mathrm{Expert}_{e}(\mathrm{Fused})\right)
  \end{gathered}
\end{equation}
where $r_{e}$ are routing logits, $w_{e}$ are gating weights, and $E$ stands for numbers of experts. In addition, $\mathrm{Expert}_{e}$ are specialized networks that capture different types of relational patterns (e.g., linear correlations, nonlinear couplings, time-delayed interactions). The output is actually the enhancement of feature $\boldsymbol{f}_{x}(i)$, ensuring the preservation of the node's intrinsic characteristics while incorporating relational insights.

\section{PolaDCA-based Message Passing}

Inspired by\cite{PolaTransformer}, we also introduce PolaDCA for GNNs, which is an advanced extension of DCA for GNNS that explicitly models the polarity of interactions between nodes. It enables the model to distinguish between synergistic (enhancing) and antagonistic (suppressing) relationships, providing a more nuanced understanding of complex interactions.

\subsection{Polarized Decomposition}

Traditional attention mechanisms, including DCA for GNNs (Eq.~(\ref{MDCA})), capture only the magnitude of interactions through non-negative attention weights. PolaDCA for GNNs extends this by learning both the magnitude and polarity of relationships, enabling explicit modeling of physical phenomena such as resonance amplification, damping effects, and compensatory behaviors. 

From node features $\boldsymbol{f}_x(i),\boldsymbol{f}_y(i)$ and $\boldsymbol{f}_z(i)$, we denote the Query, Key, and Value spaces as follows:

\begin{equation}
    \label{node_pro}
    \boldsymbol{Q}_i=\boldsymbol{f}_x(i),\quad \boldsymbol{K}_i=\boldsymbol{f}_y(i),\quad \boldsymbol{V}_i=\boldsymbol{f}_z(i)
\end{equation}

Unlike conventional attention mechanisms that operate directly on $\boldsymbol{Q}$ and $\boldsymbol{K}$, PolaDCA decomposes both Query and Key matrices into positive and negative components using rectified linear unit operations\cite{PolaTransformer}: 
\begin{equation}
  \label{eq:decomposition}
  \begin{split}
    \boldsymbol{Q}_i^{\mathrm{pos}} = \mathrm{ReLU}(\boldsymbol{Q}_i),   & \quad \boldsymbol{Q}_i^{\mathrm{neg}} = \mathrm{ReLU}(-\boldsymbol{Q}_i) \\[2pt]
    \boldsymbol{K}_i^{\mathrm{pos}} = \mathrm{ReLU}(\boldsymbol{K}_i),   &\quad \boldsymbol{K}_i^{\mathrm{neg}} = \mathrm{ReLU}(-\boldsymbol{K}_i)
  \end{split}
\end{equation}

The polarity decomposition enables the computation of four distinct attention matrices: 

(1) Positive-positive interaction (cooperative enhancement): 
\begin{equation}
    \label{eq:pp_attention}
    \boldsymbol{A}_i^{\mathrm{pp}}=\frac{{\boldsymbol{Q}}_{i}^{\mathrm{pos}}{\left ( {\boldsymbol{K}}_{i}^{\mathrm{pos}}\right )}^T }{\sqrt{{d}_{k}}}
\end{equation}
where $d_k = D / H$ is the dimension per head in multi-head attention with $H$ heads. It captures synergistic relationships where both nodes exhibit positive deviations from their states.

(2) Negative-negative interaction (shared deficiency correlation):
\begin{equation}
    \label{eq:nn_attention}
    \boldsymbol{A}_i^{\mathrm{nn}}=\frac{{\boldsymbol{Q}}_{i}^{\mathrm{neg}}{\left ( {\boldsymbol{K}}_{i}^{\mathrm{neg}}\right )}^T }{\sqrt{{d}_{k}}}
\end{equation}
This matrix models situations where nodes share common deficiencies or suppression patterns.

(3) Positive-negative interaction (compensatory relationship):
\begin{equation}
    \label{eq:pn_attention}
    \boldsymbol{A}_i^{\mathrm{pn}}=\frac{{\boldsymbol{Q}}_{i}^{\mathrm{pos}}{\left ( {\boldsymbol{K}}_{i}^{\mathrm{neg}}\right )}^T }{\sqrt{{d}_{k}}}
\end{equation}
This asymmetric interaction captures compensatory behaviors where a positive deviation in one node correlates with a negative deviation in another. 

(4) Negative-positive interaction (inverse compensatory relationship):
\begin{equation}
    \label{eq:np_attention}
    \boldsymbol{A}_i^{\mathrm{np}}=\frac{{\boldsymbol{Q}}_{i}^{\mathrm{neg}}{\left ( {\boldsymbol{K}}_{i}^{\mathrm{pos}}\right )}^T }{\sqrt{{d}_{k}}}
\end{equation}
This is the complementary case to $\boldsymbol{A}_{\mathrm{pn}}$, capturing situations where negative deviations in one node correlate with positive deviations in another.

\subsection{PolaDCA for GNNs}

Rather than employing fixed combination rules, we utilize learnable weighting parameters that allow the model to adaptively emphasize different interaction types based on the specific characteristics of nodes and fault scenario: 
\begin{equation}
    \label{com}
    \boldsymbol{A}_i^{\mathrm{com}}=w_{\mathrm{pp}}\odot \boldsymbol{A}_i^{\mathrm{pp}}+w_{\mathrm{nn}}\odot \boldsymbol{A}_i^{\mathrm{nn}}+w_{\mathrm{pn}}\odot \boldsymbol{A}_i^{\mathrm{pn}}+w_{\mathrm{np}}\odot \boldsymbol{A}_i^{\mathrm{np}}
\end{equation}
where $w_{\mathrm{pp}}, w_{\mathrm{nn}}, w_{\mathrm{pn}}, w_{\mathrm{np}} \in \mathbb{R}^{H \times 1 \times 1}$ are learnable parameters to reflect the natural tendencies of these interactions.

From Definition 2 and Eqs.~(\ref{node_pro})-(\ref{com}), we give the definition for PolaDCA for GNNs. 

\textbf{Definition 3 (PolaDCA for GNNs)}: For node $v_i$ in GNNs, given a feature set \( \mathcal{H}_i = \{\boldsymbol{f}_x(i), \boldsymbol{f}_y(i), \boldsymbol{f}_z(i)\} \) with $ \boldsymbol{W}_x^i,\boldsymbol{W}_y^i,\boldsymbol{W}_z^i$ being projection matrices.
The PolaDCA for GNNs is defined by 

\begin{equation}
    \text{PolaDCA}(\boldsymbol{f}_x(i), \boldsymbol{f}_y(i), \boldsymbol{f}_z(i))  
     =\mathrm{softmax}(\boldsymbol{A}_i^{\mathrm{com}})\boldsymbol{V}_i 
    \label{PolaDCA}
\end{equation}
where \( \boldsymbol{W}_x^i, \boldsymbol{W}_y^i, \boldsymbol{W}_z^i \), $w_{\mathrm{pp}}, w_{\mathrm{nn}}, w_{\mathrm{pn}}$ and $w_{\mathrm{np}}$ are learnable projection matrices. 

Thus, the final output of PolaDCA for GNNs is given by
\begin{equation}
  \label{eq:final_output}
  \begin{gathered}
    \mathrm{Output}=\sigma(\boldsymbol{W}_o\cdot\mathrm{PolaDCA}(\boldsymbol{f}_x(i),\boldsymbol{f}_y(i),\boldsymbol{f}_z(i)))
  \end{gathered}
\end{equation}
where $\boldsymbol{W}_o \in \mathbb{R}^{D \times D}$ is the output projection matrix. 
The complete computational flow of PolaDCA for GNNs is illustrated in Fig.~\ref{PolaDCA}. PolaDCA represents a significant advancement in relational reasoning for GNNs.

\begin{figure*}[t]       
  \centering
  \includegraphics[width=4.5in]{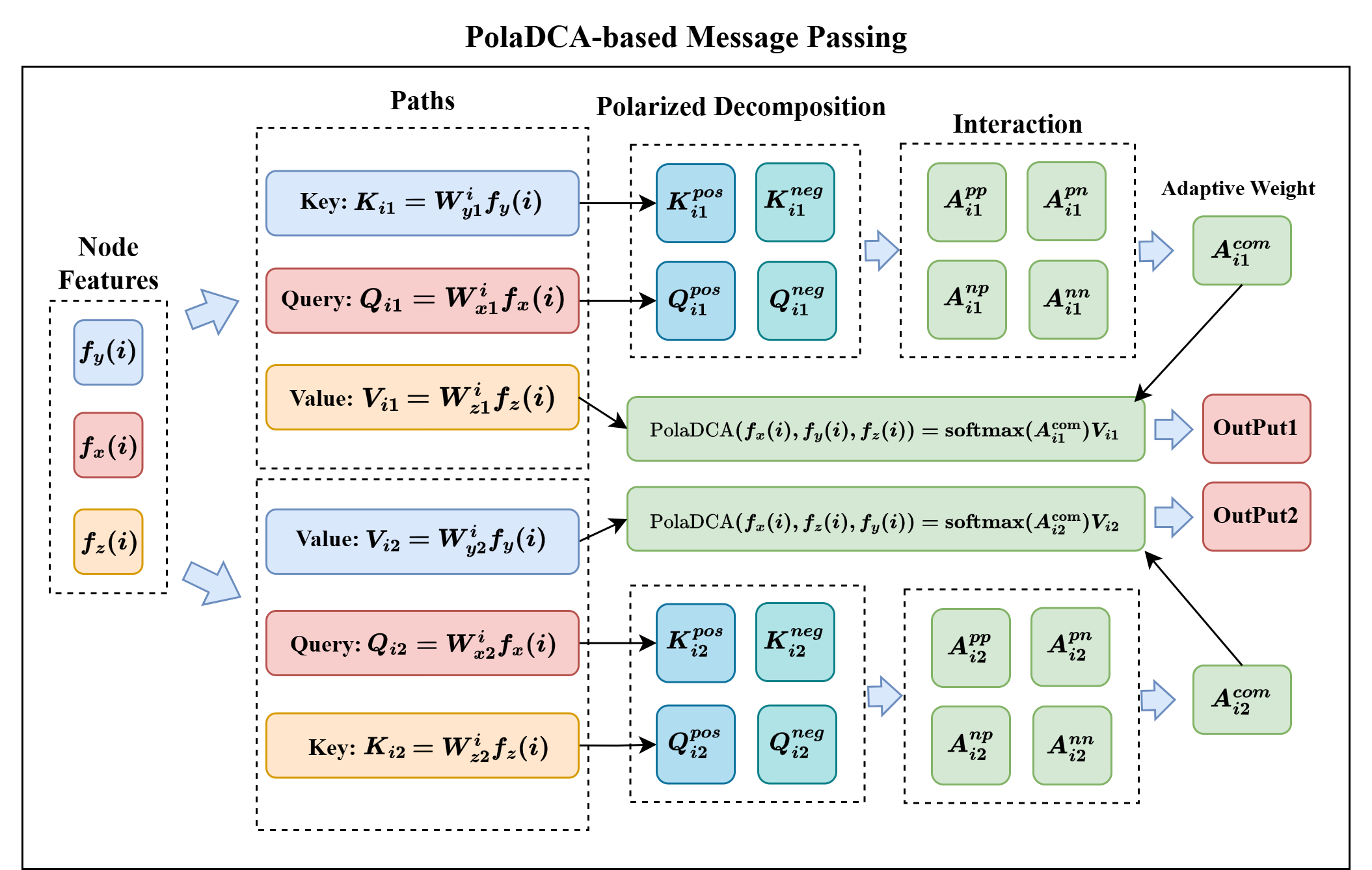}
  \caption{Framework of PolaDCA-based message passing}
  \label{PolaDCA}
\end{figure*}

To evaluate algorithmic efficiency, we quantify the computational complexity of PolaDCA-GNN against SCA and DCA-GNN based on per-layer floating-point operations (FLOPs). While the SCA maintains a lower complexity of $2n^2 D$
focusing on spatial attention scores, the DCA-GNN introduces feature projections that elevate the cost to $2nD^2 + 2n^2 D$. 
Notably, PolaDCA-GNN results in a total complexity of $2nD^2 + 2n^2 D + 4nD$. Despite the inclusion of polarization operators, PolaDCA-GNN preserves the same asymptotic growth rate as DCA-GNN, ensuring its high-performance diagnostic capabilities are computationally viable for real-time industrial monitoring.

\subsection{Loss Function}

A combination of the log-softmax transformation and the negative log-likelihood (NLL) loss can be used to train DCA/PolaDCA for GNNs. It is mathematically equivalent to the loss of cross-entropy but offers superior numerical stability\cite{CE-Loss}. The loss function $\mathcal{L}$ for a single sample is defined as
\begin{equation}
\label{LossFuction}
    \mathcal{L}(\mathbf{z}, y) = -\log \left( \frac{\exp(z_y)}{\sum_{j=1}^{C} \exp(z_j)} \right) = -z_y + \log \sum_{j=1}^{C} \exp(z_j)
\end{equation}
where $y \in \{1, \dots, K\}$ represents the index of the true label. The Adam optimizer is used to minimize $\mathcal{L}$ and update trainable parameters $\boldsymbol{\Theta}$ \cite{Adam}. 

\section{Performance analysis on noise robustness}

\subsection{Model Definitions (Single-layer)}

The function \( f_{\mathrm{GCN}}(\cdot) \) denotes the single-layer transformation of a GCN, which updates the node representations through normalized neighborhood aggregation. The output of each node is computed by averaging the features of itself and its neighbors, scaled by node degrees, followed by a linear transformation and activation (Eqs.~(\ref{layer-propa})-(\ref{GCNs})).

The function \( f_{\mathrm{DCA}}(\cdot) \) represents the single-layer transformation of a DCA for GNNs (Eq.~(\ref{MDCA})). Three features per node can be dynamically computed: the projected node feature \( f_x(i) \), the neighborhood consensus \( f_y(i) \), and the neighborhood diversity \( f_z(i) \) (Eqs.~(\ref{X})-(\ref{diversity})). These interact through a content-based attention mechanism that correlates the node feature with consensus, weighted by diversity (Eq.~(\ref{MDCA})).

The function \( f_{\mathrm{PolaDCA}}(\cdot) \) denotes the single-layer transformation of a PolaDCA module. It maps the features of input nodes to updated representations by first projecting them into Query, Key, and Value spaces (Eq.~(16)), then decomposing Query and Key into positive/negative components via ReLU (Eq.~(17)). Four attention matrices are computed (Eqs.~(18)-(21)), adaptively fused with learnable signed weights (Eq.~(22)), and finally normalized and projected (Eq.~(23)).

\subsection{Assumptions and Lemmas}

\textbf{Assumption 1 }(Lipschitz continuity): 
    
(1) Activation \(\sigma\) is \(L_\sigma\)-Lipschitz;

(2) Weight matrices satisfy \(\| \boldsymbol{W} \|_2 \leq B_W\);

(3) Input features satisfy \(\| \boldsymbol{x}_i \| \leq B_x\); 

(4) Softmax is 1-Lipschitz w.r.t. infinity norm.

\textbf{Assumption 2} (Noise model): 
The noisy feature matrix is modeled as \(\boldsymbol{X}_\eta = \boldsymbol{X} + \boldsymbol{\eta}\), where \(\boldsymbol{\eta} \in \mathbb{R}^{N \times D}\) is additive noise with \(\eta_{ij} \sim \mathcal{N}(0, \bar{\sigma}^2)\), \(i = 1, \ldots, N\), \(j = 1, \ldots, D\). For correlated noise scenarios, we have $
\mathbb{E}[\boldsymbol{\eta}_i \boldsymbol{\eta}_j] = \rho_{ij} \bar{\sigma}^2,\, \rho_{ij} \in [-1, 1]$, 
where \(\boldsymbol{\eta}_i, \boldsymbol{\eta}_j\) denotes noise vectors for nodes \(i\) and \(j\).

\textbf{Assumption 3} (Feature extraction): 
The mapping \(g: \mathbb{R}^L \to \mathbb{R}^D\) from raw signals to node features is \(L_g\)-Lipschitz, with \(L_g \leq C / \sqrt{L}\) for window length \(L\).

\textbf{Assumption 4} (Optimal learning): Models are trained to optimality or near-optimality:
    
(1) DCA for GNNs (Eq.~(\ref{MDCA})) learns attention weights that minimize both task loss and noise sensitivity; 

(2) PolaDCA for GNNs (Eq.~(\ref{PolaDCA})) learns signed weights that maximize both task performance and noise cancellation; 

(3) Gradient-based optimization converges to solutions that meet these properties.

Lemmas 1-3 are necessary to prove the main theorems. 

\textbf{Lemma 1} (Consensus feature Lipschitz): 
For the consensus feature defined in Eq.~(\ref{consensus}), an upper bound on its sensitivity to input perturbations is given by we have
\begin{equation}
   \| \boldsymbol{f}_{y}(i)(\boldsymbol{X} + \boldsymbol{\eta}) - \boldsymbol{f}_{y}(i)(\boldsymbol{X})\| \leq B_{W}\cdot \frac{1}{|\mathcal{N}(i)|}\sum_{j\in \mathcal{N}(i)}\| \boldsymbol{\eta}_{j}\|
   \label{CF-Lipschitz}
\end{equation}

The detailed proof is given in the Appendix~\ref{app:lemma1}.

\textbf{Lemma 2} (Diversity feature Lipschitz): For the diversity feature in Eq.~(\ref{diversity}), we have 
\begin{equation}
  \| \boldsymbol{f}_{z}(i)(\boldsymbol{X} + \boldsymbol{\eta}) - \boldsymbol{f}_{z}(i)(\boldsymbol{X})\| \leq \sqrt{2} B_{W}\cdot \sqrt{\frac{1}{|\mathcal{N}(i)|}\sum_{j\in \mathcal{N}(i)}\| \boldsymbol{\eta}_{j}\|^{2}}
  \label{DF-Lipschitz}
\end{equation}

The detailed proof is given in the Appendix~\ref{app:lemma2}.

Based on Lemmas 1–2, we further analyze the Lipschitz continuity of the DCA operator for GNNs (Eq.~(\ref{MDCA})). 

\textbf{Lemma 3}: For the DCA operator (Eq.~(\ref{MDCA})), suppose that \(\| \boldsymbol{a}\| ,\| \boldsymbol{a}'\|,\| \boldsymbol{b}\| ,\| \boldsymbol{b}'\|,\| \boldsymbol{c}\|,\| \boldsymbol{c}'\| \leq M\). Then, the following inequality holds:

\begin{equation}
\begin{split}
&\| \mathrm{DCA}(\boldsymbol{a}',\boldsymbol{b}',\boldsymbol{c}') - \mathrm{DCA}(\boldsymbol{a},\boldsymbol{b},\boldsymbol{c})\| \\
&\leq \frac{1}{\sqrt{d_k}} (\| \boldsymbol{a}'\| \| \boldsymbol{b}' - \boldsymbol{b}\| +\| \boldsymbol{b}\| \| \boldsymbol{a}' - \boldsymbol{a}\|) + \| \boldsymbol{c}' - \boldsymbol{c}\|
\label{Delta:error}
\end{split}
\end{equation}

The detailed proof is given in the Appendix~\ref{app:lemma3}.

\subsection{Main Theorems on Noise Robustness}
Based on the Lipschitz properties established in Lemmas 1–3 and in Assumptions 1–4, we now compare the noise robustness of PolaDCA and GCN for GNNs.

\textbf{Theorem 1} (PolaDCA vs GCN noise robustness): There exist constants \(C_1, C_2 > 0\) such that
\begin{equation}
    \mathbb{E}\| f_{\mathrm{PolaDCA}}(\boldsymbol{X}_{\eta}) - f_{\mathrm{PolaDCA}}(\boldsymbol{X})\|_F \leq C_1\bar{\sigma},
\end{equation}
\begin{equation}
    \mathbb{E}\| f_{\mathrm{GCN}}(\boldsymbol{X}_{\eta}) - f_{\mathrm{GCN}}(\boldsymbol{X})\|_F \leq C_2\bar{\sigma},
\end{equation}
and \(C_1 < C_2\).

The proof leverages the layer-wise perturbation analysis and the cancellation effect introduced by polarized attention mechanism of PolaDCA (see Appendix~\ref{app:theorem1} for details).

Extending the analysis to DCA for GNNs, we show that DCA also exhibits stronger noise robustness than GCN under the same assumptions. 

\textbf{Theorem 2} (DCA vs GCN noise robustness): 
There exist \(L_{\mathrm{DCA}}, L_{\mathrm{GCN}} > 0\) such that
\begin{equation}
    \mathbb{E}\| f_{\mathrm{DCA}}(\boldsymbol{X}_{\eta}) - f_{\mathrm{DCA}}(\boldsymbol{X}) \|_F \leq L_{\mathrm{DCA}} \bar{\sigma},
\end{equation}
\begin{equation}
    \mathbb{E}\| f_{\mathrm{GCN}}(\boldsymbol{X}_{\eta}) - f_{\mathrm{GCN}}(\boldsymbol{X}) \|_F \leq L_{\mathrm{GCN}} \bar{\sigma},
\end{equation}
and \(L_{\mathrm{DCA}} < L_{\mathrm{GCN}}\).

The proof exploits the learnable attenuation factor \(\epsilon > 0\) enabled by DCA’s adaptive projections (see Appendix~\ref{app:theorem2}).

When noise exhibits heterogeneous correlation structures, PolaDCA demonstrates further improvement over DCA.

\textbf{Theorem 3} (PolaDCA vs DCA with heterogeneous noise): 
If noise has a heterogeneous correlation structure, then there exist \(L_{\mathrm{DCA}}, L_{\mathrm{PolaDCA}} > 0\) such that
\begin{equation}
   \mathbb{E}\| f_{\mathrm{PolaDCA}}(\boldsymbol{X}_\eta) - f_{\mathrm{PolaDCA}}(\boldsymbol{X}) \|_F \leq L_{\mathrm{PolaDCA}}\bar{\sigma},
\end{equation}
and \(L_{\mathrm{PolaDCA}} < L_{\mathrm{DCA}}\).

The reduction comes from the ability of PolaDCA to assign signed attention weights, thereby exploiting negative correlations for noise cancellation (complete proof in Appendix~\ref{app:theorem3}).

Combining Theorems 1–3, we establish a strict hierarchy of noise robustness among the three architectures.

\textbf{Theorem 4 }(Noise robustness hierarchy): 
For signals with additive noise satisfying Assumptions 1-4, the optimal Lipschitz constants satisfy
\begin{equation}
    L_{\mathrm{PolaDCA}}^* < L_{\mathrm{DCA}}^* < L_{\mathrm{GCN}}^*.
\end{equation}

This result summarizes the theoretical advantage of PolaDCA for GNNs (Eq.~(\ref{PolaDCA})), followed by DCA for GNNs (Eq.~(\ref{MDCA})), over the baseline GCN in terms of noise stability. The proof follows from the progressive reduction of Lipschitz constants through architectural innovations (see Appendix~\ref{app:theorem4}). 

\section{PolaDCA-based fault diagnosis}

The off-training and online diagnosis of PolaDCA-GNN are implemented by Algorithms 1 and 2, respectively. The former determines the confidence thresholds through kernel density estimation (KDE) \cite{KDE_S} and Monte carlo dropout technique \cite{Dropout}. The latter applies the trained model to the testing data. This incorporates uncertainty-aware decision making.
The diagnostic decision is made based on both the predicted fault class and the estimated uncertainty levels.

Let $\boldsymbol{Z}_t \in \mathbb{R}^d$ denote the latent representation enhanced by PolaDCA and dynamic expert fusion at time step $t$. The decision module employs a multi-layer perceptron (MLP) 
as a non-linear projector $G: \mathbb{R}^d \to \mathbb{R}^K$. The transformation can be formulated as $
    \boldsymbol{P}_{logits} = \boldsymbol{W}_L \cdot \sigma(\dots \sigma(\boldsymbol{W}_1 \boldsymbol{Z}_t + \boldsymbol{b}_1) \dots) + \boldsymbol{b}_L$. 
To quantify the uncertainty of the diagnosis, a softmax layer is applied to the logits to generate a posterior probability distribution. The final predicted fault status $\hat{y}$ is determined using the Maximum A Posteriori (MAP) criterion \cite{CE-Loss}:

\begin{equation}
\label{id}
    \hat{y} = \text{arg max}_{k \in \{1, \dots, K\}} \frac{\exp(P_{logits}^{(k)})}{\sum_{j=1}^{K} \exp(P_{logits}^{(j)})}
\end{equation}

\begin{algorithm}[t]
\caption{PolaDCA-GNN: Offline Training}
\label{alg:training}
\begin{algorithmic}[1]
\renewcommand{\algorithmicrequire}{\textbf{Input:}}
\renewcommand{\algorithmicensure}{\textbf{Output:}}
\REQUIRE Training set $\mathcal{D}_{train}$, iterations $I$, expert count $E=3$, neighbors $K=8$.
\ENSURE Optimized model parameters ${\theta}^*$.

\STATE \textbf{Initialization:} Randomly initialize weights ${\theta}$ and construct $k$-NN topology $\boldsymbol{A}$.
\FOR{$epoch = 1$ to $I$}
    \FOR{each batch $(\mathcal{B}_{\boldsymbol{X}}, \mathcal{B}_y) \subset \mathcal{D}_{train}$}
        \STATE \textbf{Message passing:} Aggregate features $f_x(i)$, $f_y(i)$, $f_z(i)$ by Eqs.~(\ref{X})-(\ref{diversity});
        \STATE \textbf{Polarized decomposition} via Eqs.~(\ref{node_pro})-(\ref{eq:np_attention});
        \STATE \textbf{Interaction:} Compute polarized attention scores using $\{w_{pp}, w_{nn}, w_{mn}, w_{mp}\}$ by Eqs.~(\ref{com});
        \STATE \textbf{Dynamic expert routing:} Calculate gating weights $w_{e}$ and enhancement factor $\alpha$ via Eqs.~(\ref{path1})-(\ref{deqn_exla});
        \STATE \textbf{Objective optimization:} Compute cross-entropy loss $\mathcal{L}$ via Eq.~(\ref{LossFuction}) and update ${\boldsymbol{\Theta}}$ via backpropagation;
    \ENDFOR
    \STATE \textbf{Learning rate decay:} $\eta \leftarrow \eta \times \gamma$ at scheduled intervals.
\ENDFOR
\RETURN ${\theta}^*$
\end{algorithmic}
\end{algorithm}

\begin{algorithm}[t]
\caption{PolaDCA-GNN: Online Diagnosis}
\label{alg:diagnosis}
\begin{algorithmic}[1]
\REQUIRE Real-time stream $\mathcal{S}(t)$, trained $\mathcal{M}({\theta}^*)$, window $(T, s)$.
\ENSURE Fault category $\hat{y}$ and routing weights $w_{e}$.
\STATE Load pre-trained parameters ${\theta}^*$.
\WHILE{industrial process is active}
    \STATE \textbf{Sampling:} Extract current segment $\boldsymbol{W}_t$ from $\mathcal{S}(t)$;
    \STATE \textbf{Dynamic mapping:} Update adjacency $\boldsymbol{A}_t$ based on instantaneous correlations;
    \STATE \textbf{Inference:}
    \STATE \quad 1) Execute PolaDCA via Eqs.~(\ref{node_pro})-(\ref{eq:final_output});
    \STATE \quad 2) Execute dynamic expert fusion to compute gating scores $w_{e}$ via Eqs.~(\ref{path1})-(\ref{deqn_exla});
    \STATE \textbf{Decision:} Identify fault status $\hat{y}$ via Eq.~(\ref{id});
    \STATE \textbf{Slide Window:} $t \leftarrow t + s$.
\ENDWHILE
\end{algorithmic}
\end{algorithm}

\section{Experimental Study}
\subsection{Dataset Description and Preprocessing}
To validate the proposed model, we employ three real-world datasets for fault diagnosis: the XJTUSuprgear \cite{XJTU}, the CWRUBearing \cite{CWRU_Dataset,CWRU_Use} and the Three-Phase Flow Facility (TFF) datasets \cite{TTF}. Each datasets are preprocessed into a graph-structured format for subsequent analysis. Each dataset is split 7:3 into training and testing sets. The detailed description of the dataset is given in Appendix~\ref{app:datasets}. 

The XJTUSuprgear dataset contains vibration and acoustic emission signals from a gear test rig under ten operating conditions. Raw time-series are segmented via a sliding window of length 1,000 and stride 500. A repetition sampling rate of 0.5 is applied, yielding 200 samples per condition. A graph is constructed per sample using $k$-nearest neighbors \cite{kNN} ($k=8$). The sensor position and fault types in spur gear can be inferred in Fig.~\ref{fig:xjtu_sensors} and Table~\ref{tab:xjtu_fault_types_app}, respectively. 

The CWRUBearing dataset (Drive End accelerometer) includes ten states under load 0, 48 kHz sampling. Signals are segmented with a window of 1,000 and stride 64, and a repetition sampling rate of 0.936 is used to generate 200 samples per condition. Here, $k=20$ is chosen for the construction of the nearest neighbor graph. The bearing fault types can be inferred in Table~\ref{tab:cwru_fault_types_app}.

The TFF dataset consists of sensor signals collected from an industrial multiphase flow facility under six operating conditions. Raw time-series data are pre-processed using $Z$-score standardization and segmented via a sliding window of length 100 and stride 50. For each windowed sample, a graph is constructed where 100 time steps serve as nodes, with connectivity established through the nearest neighbors $k=8$. 
The variables and fault types in TFF can be depicted in Table~\ref{table_variables_app} and Table~\ref{tab:ttf_fault_types_app}, respectively.

\subsection{Experimental Settings}

To evaluate the effectiveness of DCA/PolaDCA, seven state-of-the-art graph-based fault diagnosis methods are selected as baselines for comparison: GCN \cite{GCN}, GAT \cite{GAT}, GraphSAGE \cite{GraphSAGE}, GTF \cite{GTF}, GCL \cite{GCL_main}, MRF-GCN \cite{MRFGNN}, IAGNN \cite{IAGNN}, FIGNN \cite{FIGNN}, CDGNN \cite{CDGNN} and MA-STGNN \cite{MA-STGNN}. 

The DCA-/PolaDCA-GNN models are configured with the following architecture: input projection layer to transform raw features to a hidden dimension of \(d_{\text{model}} = 64\), followed by three graph layers with \(n_{\text{head}} = 4\) attention heads per layer. Each layer uses a dropout rate of \(0.01\) for regularization. The classifier consists of two fully connected layers with dimensions 128 and 64, respectively, and the output layer corresponding to the number of fault classes (10 for XJTUSuprgear and CWRUBearing datasets, and 6 for TTF dataset).

The model is trained for 50 epochs using the Adam optimizer with a learning rate of \(0.001\), weight decay of \(5 \times 10^{-4}\), and batch size of 16. A step learning rate scheduler is applied with a decay factor of \(0.5\) every 20 epochs. Early stopping is used if the validation loss does not improve for 20 consecutive epochs. All models are trained and evaluated on the same fixed train-test split (7:3) to ensure fair comparison. In addition, all experiments were conducted with 50 epochs, and each dataset was randomly tested five times independently. The mean and standard deviation of the accuracy and the F1-score were calculated as evaluation metrics.

\subsection{Fault Diagnosis Performances}

The experimental results are presented in Tables~\ref{XJTUN}-\ref{TFF_Normal}, which compare the performance of DCA-/PolaDCA-GNN against seven state-of-the-art baseline methods in terms of both accuracy (ACC) and macro-F1 score.

\subsubsection{XJTUSuprgear Dataset}

As illustrated in Table~\ref{XJTUN}, DCA-/PolaDCA-GNN demonstrate superior diagnostic capabilities on the XJTUSuprgear dataset compared to other methods. Under normal operating conditions, PolaDCA-GNN achieves a perfect accuracy of $100\%$ in its best test and $99.53\%$ both accuracy and F1-socre on average in only 50 epochs, significantly outperforming traditional models like GCN ($94.97\%$), GAT ($78.97\%$), GTF ($97.99\%$). Although GTF uses the SCA for messaging passing, its performance is much worse than DCA-/PolaDCA-GNNs. This partially shows the superiority of DCA/PolaDCA over SCA.  

To further validate the robustness (Table II), Gaussian noise are injected into raw signals with signal-to-noise ratios (SNR) ranging from $0\,\text{dB}$ to $-8\,\text{dB}$. In strong cases (down to –8\,dB SNR), DCA‑/PolaDCA‑GNN consistently outperform all baseline models (GCN, GAT, etc.), maintaining the highest accuracy and F1‑scores. DCA‑GNN shows particularly robust performance at the most extreme noise levels (–6\,dB and –8\,dB), slightly surpassing PolaDCA‑GNN, while PolaDCA‑GNN leads at milder noise levels (0\,dB to –4\,dB).

\begin{table}[t!]
\centering
\caption{Experimental Result on XJTUSuprgear Normal Data}
\label{tab:results}
\begin{tabular}{lcccc}
\toprule
\multirow{2}{*}{\textbf{Method}} & \multicolumn{2}{c}{\textbf{Average (\%)}} & \multicolumn{2}{c}{\textbf{Best (\%)}} \\
\cmidrule(lr){2-3} \cmidrule(lr){4-5}
 & \textbf{ACC} & \textbf{F1} & \textbf{ACC} & \textbf{F1} \\
\midrule
GCN            & $94.97\pm3.37$ & $93.69\pm2.97$ & 96.67 & 96.67 \rule[0pt]{0pt}{8pt}\\  
GAT            & $78.97\pm6.29$ & $77.70\pm7.07$ & 89.17 & 89.06 \rule[0pt]{0pt}{8pt}\\
GraphSAGE      & $87.57\pm1.38$ & $87.46\pm1.48$ & 88.83 & 88.74 \rule[0pt]{0pt}{8pt}\\
GTF(SCA)            & $97.99\pm1.00$ & $98.00\pm0.99$ & 99.00 & 99.00 \rule[0pt]{0pt}{8pt}\\
GCL            & $81.80\pm2.95$ & $81.48\pm3.11$ & 85.00 & 84.82 \rule[0pt]{0pt}{8pt}\\
MRF-GCN        & $84.98\pm0.62$ & $84.97\pm0.65$ & 85.83 & 85.88 \rule[0pt]{0pt}{8pt}\\
IAGNN          & $90.97\pm0.95$ & $90.93\pm0.96$ & 91.83 & 91.77 \rule[0pt]{0pt}{8pt}\\
\textbf{DCA-GNN}  & $99.32\pm0.50$ & $99.32\pm0.50$ & \textbf{100} & \textbf{100} \rule[0pt]{0pt}{8pt}\\
\textbf{PolaDCA-GNN} & $\mathbf{99.53\pm0.30}$ & $\mathbf{99.53\pm0.30}$ & \textbf{100} & \textbf{100} \rule[0pt]{0pt}{8pt}\\
\bottomrule
\end{tabular}
\label{XJTUN}
\end{table}

Under normal data, the complete confusion matrices on all contrastive methods in Fig.~\ref{fig:CMXJTU_app}. For PolaDCA, the evolution of adaptive attention weights in Fig.~\ref{fig:Weight_I_app}.  
From this figure, positive–positive cooperation ($w_{pp}$) remains the most influential interaction, followed by negative–negative suppression ($w_{nn}$), while two types of compensatory interaction ($w_{pn}$ and $w_{np}$) play relatively weaker roles. This demonstrates PolaDCA's capability to dynamically model nuanced node relationships and prioritize interaction types.
The t-SNE visualization in Fig.~\ref{fig:tSNE_app} qualitatively confirms that PolaDCA-GNN produces more compact intra-class clusters and wider inter-class margins. This underscores the superior feature extraction and noise-rejection capabilities of PolaDCA.
While PolaDCA employs globally universal attention weights, their response to different working conditions is heterogeneous. In Fig~\ref{fig:XJTUdiff_app}, $w_{pp}$ and $w_{nn}$ are prominently dominant in healthy states, while fault-induced data trigger a distinct redistribution of weight importance.

\renewcommand{\arraystretch}{1.2}
\setlength{\tabcolsep}{2.2pt} 
\newcolumntype{C}[1]{>{\centering\arraybackslash}p{#1}}
\begin{table*}[t!]
\centering
\caption{Average Experimental Result on XJTUSuprgear with Gaussian Noise}
\label{tab:cross}
\begin{tabular}{l *{10}{C{1.15cm}}}
\toprule
\multirow{2.5}{*}{\textbf{Method}} & 
\multicolumn{2}{c}{\textbf{0\,dB}} &
\multicolumn{2}{c}{\textbf{-2\,dB}} &
\multicolumn{2}{c}{\textbf{-4\,dB}} &
\multicolumn{2}{c}{\textbf{-6\,dB}} &
\multicolumn{2}{c}{\textbf{-8\,dB}} \\
\cmidrule(lr){2-3} \cmidrule(lr){4-5} \cmidrule(lr){6-7} \cmidrule(lr){8-9} \cmidrule(lr){10-11}
 & \textbf{ACC(\%)} & \textbf{F1(\%)} &
   \textbf{ACC(\%)} & \textbf{F1(\%)} &
   \textbf{ACC(\%)} & \textbf{F1(\%)} &
   \textbf{ACC(\%)} & \textbf{F1(\%)} &
   \textbf{ACC(\%)} & \textbf{F1(\%)} \\
\midrule
GCN       & 75.17 & 74.57 & 71
83 & 71.15 & 71.77 & 71.06 & 70.50 & 69.92 & 62.17 & 61.13 \\
GAT       & 62.17 & 55.94 & 60.17 & 54.39 & 61.33 & 57.89 & 60.67 & 57.67 & 46.67 & 37.37 \\
GraphSAGE & 69.33 & 69.51 & 67.83 & 67.70 & 61.89 & 61.85 & 59.17 & 58.06 & 60.33 & 60.70 \\
GTF(SCA)       & 84.22 & 84.11 & 83.83 & 83.71 & 78.67 & 78.45 & 79.17 & 78.65 & 77.06 & 76.58 \\
GCL       & 49.56 & 46.71 & 47.33 & 44.44 & 43.56 & 39.95 & 45.00 & 40.82 & 43.67 & 39.53 \\
MRF-GCN   & 54.50 & 53.86 & 51.00 & 50.91 & 45.67 & 44.83 & 41.67 & 41.33 & 35.00 & 33.60 \\
IAGNN     & 70.00 & 69.39 & 69.83 & 69.83 & 70 & 69.31 & 58 & 56.78 & 49.17 & 47.05 \\
\textbf{DCA-GNN}     & 88.28 & 88.07 & 87.17 & 87.01 & 85.21 & 84.83 & \textbf{81.50} & \textbf{80.99} & \textbf{80.06} & \textbf{79.17} \\
\textbf{PolaDCA-GNN} &
\textbf{89.33} & \textbf{89.21} &
\textbf{87.69} & \textbf{87.50} &
\textbf{85.33} & \textbf{84.92} &
80.83 & 80.05 &
79.33 & 78.82 \\
\bottomrule
\end{tabular}
\label{XJTU_GN}
\end{table*}

\subsubsection{CWRUBearing Dataset}

\begin{figure}[t]       
\centering
\includegraphics[width=3in]{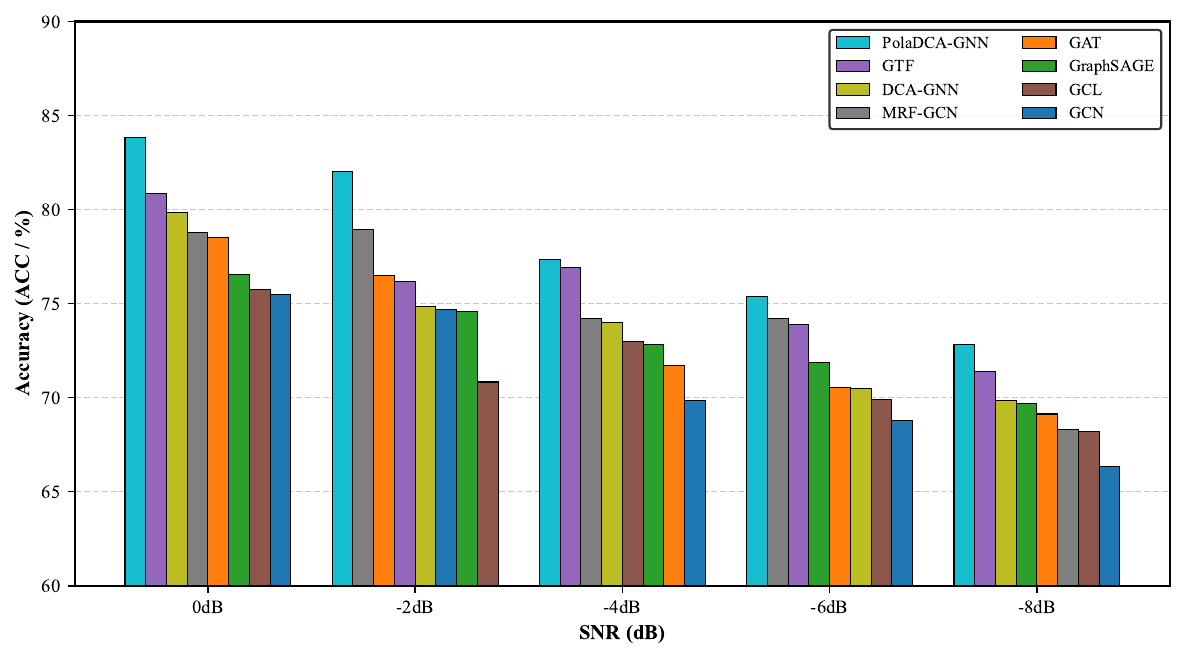}
\caption{Accuracy comparison on CWRUBering data with gaussian noise}
\label{ACC_Compari_CWRU_Noise}
\end{figure}

The comparative results on the CWRUBearing dataset are presented in Table~\ref{CWRU_Normal}. The CWRU dataset poses a challenge for models like CDGNN ($92.40\%$), GTF($96.00\%$) and GAT ($83.33\%$) with a limited number of epochs, even the best accuracy of these comparative methods can only reach $98.33\%$. In contrast, PolaDCA-GNN achieves a state-of-the-art average accuracy of $98.96\%$ and $100\%$ in its best trial. Obviously, DCA-/PolaDCA-GNN outporforms GTF with SCA being the message passing scheme. Fig.~\ref{ACC_Compari_CWRU_Noise} demonstrates the accuracy trends under varying levels of Gaussian noise. The precision of most baseline methods is observed to fluctuate or drop significantly below $80\%$ when the SNR reaches $-6\,\text{dB}$. However, the proposed models exhibit a more stable performance curve, maintaining high precision across all noise levels. This stability can be attributed to the dynamic gating mechanism and polarized feature fusion, which allow the network to adaptively ignore noisy substructures. 

\begin{table}[t!]
\centering
\caption{Experimental Result on CWRUBearing Normal Data}
\label{tab:results}
\begin{tabular}{lcccc}
\toprule
\multirow{2}{*}{\textbf{Method}} & \multicolumn{2}{c}{\textbf{Average (\%)}} & \multicolumn{2}{c}{\textbf{Best (\%)}} \\
\cmidrule(lr){2-3} \cmidrule(lr){4-5}
 & \textbf{ACC} & \textbf{F1} & \textbf{ACC} & \textbf{F1} \\
\midrule
GNN            & $95.46\pm2.18$ & $94.27\pm2.06$ & 96.91 & 96.48 \rule[0pt]{0pt}{8pt}\\  
GAT            & $83.33\pm3.67$ & $82.34\pm3.07$ & 86.82 & 86.31 \rule[0pt]{0pt}{8pt}\\
GraphSAGE      & $97.17\pm0.97$ & $97.16\pm0.87$ & 98.12 & 98.12 \rule[0pt]{0pt}{8pt}\\
GTF(SCA)            & $96.00\pm0.57$ & $96.00\pm0.41$ & 96.85 & 96.76 \rule[0pt]{0pt}{8pt}\\
GCL            & $81.17\pm1.19$ & $79.07\pm0.96$ & 82.08 & 80.15 \rule[0pt]{0pt}{8pt}\\
MRF-GCN        & $98.06\pm0.55$ & $98.01\pm0.45$ & 98.33 & 98.33 \rule[0pt]{0pt}{8pt}\\
CDGNN          & $92.40\pm0.65$ & $92.17\pm0.87$ & 93.50 & 93.48 \rule[0pt]{0pt}{8pt}\\
\textbf{DCA-GNN}          & $96.58\pm0.63$ & $95.48\pm0.34$ & 96.99 & 96.04 \rule[0pt]{0pt}{8pt}\\
\textbf{PolaDCA-GNN} & $\mathbf{98.96\pm1.05}$ & $\mathbf{98.85\pm0.95}$ & \textbf{100} & \textbf{100} \rule[0pt]{0pt}{8pt}\\
\bottomrule
\end{tabular}
\label{CWRU_Normal}
\end{table}

\subsubsection{Three-Phase Flow Facility (TFF) Dataset}

\begin{figure}[t]       
\centering
\includegraphics[width=3in]{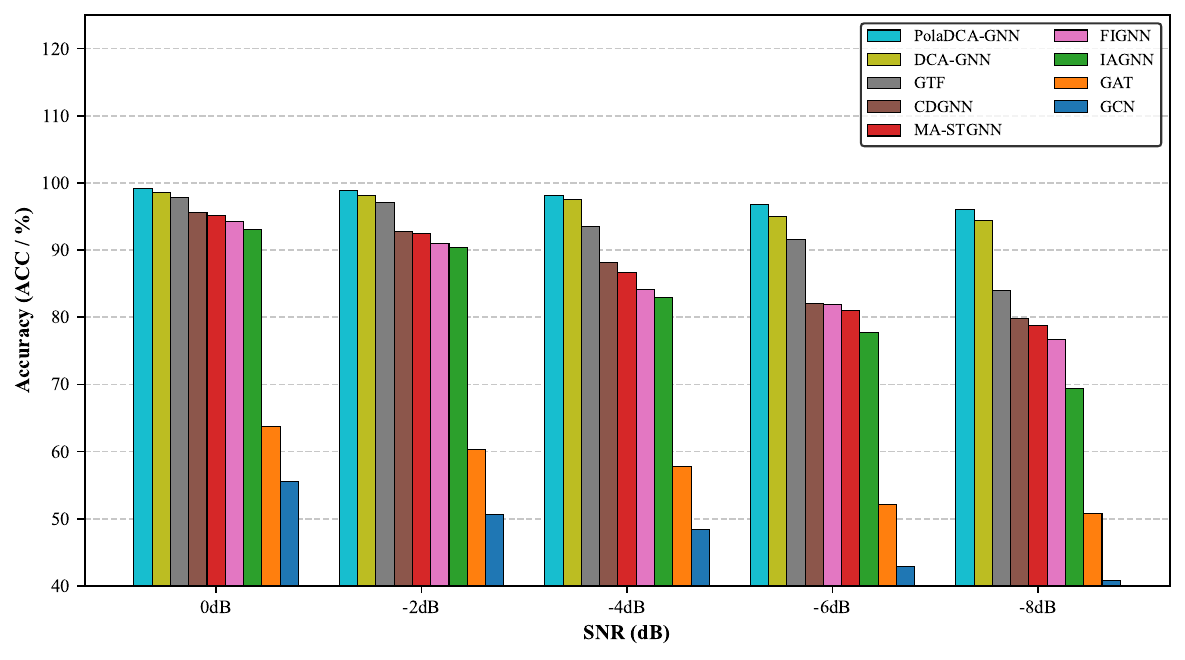}
\caption{Accuracy comparison on TFF data with gaussian noise}
\label{ACC_Compari_TFF_Noise}
\end{figure}

The TFF dataset represents an industrial process with highly coupled spatiotemporal variables. As shown in Table~\ref{TFF_Normal}, the TFF data is difficult for conventional GNNs, with GCN only attaining an average accuracy of $67.67\%$ with limited training time. Advanced models like MA-STGNN and FIGNN achieve approximately $92.80\%$ and $93.97\%$, respectively.
Nevertheless, PolaDCA-GNN outperforms all counterparts including GTF (choosing SCA as the message passing), with a leading accuracy of $99.47\%$.
This verifies the model's robustness and broad applicability across diverse sensor data types for fault diagnosis, rather than being limited to a single data type. Fig.~\ref{ACC_Compari_TFF_Noise} shows the accuracy comparison under gaussian noise, the proposed method contains a highly accuracy compared with other baseline methods. DCA-/PolaDCA-GNN maintain an accuracy of over $90\%$ even under -8db SNR gaussian noise, while other methods could only get an accuracy below $85\%$, which reflects the excellent noise robustness of DCA-/PolaDCA-GNN on this dataset.

\begin{table}[t!]
\centering
\caption{Experimental Result on TFF Noraml Data}
\label{tab:results}
\begin{tabular}{lcccc}
\toprule
\multirow{2}{*}{\textbf{Method}} & \multicolumn{2}{c}{\textbf{Average (\%)}} & \multicolumn{2}{c}{\textbf{Best (\%)}} \\
\cmidrule(lr){2-3} \cmidrule(lr){4-5}
 & \textbf{ACC} & \textbf{F1} & \textbf{ACC} & \textbf{F1} \\
\midrule
GCN            & $67.67\pm4.37$ & $65.85\pm3.97$ & 72.08 & 71.65 \rule[0pt]{0pt}{8pt}\\  
GAT            & $77.97\pm2.29$ & $76.70\pm2.07$ & 81.09 & 80.84 \rule[0pt]{0pt}{8pt}\\
IAGNN            & $91.19\pm1.65$ & $90.67\pm1.58$ & 93.45 & 92.17 \rule[0pt]{0pt}{8pt}\\
MA-STGNN            & $93.80\pm2.86$ & $92.78\pm2.57$ & 96.85 & 95.66 \rule[0pt]{0pt}{8pt}\\
CDGNN        & $97.58\pm0.83$ & $96.97\pm0.71$ & 98.14 & 97.87 \rule[0pt]{0pt}{8pt}\\
FIGNN          & $95.97\pm0.72$ & $95.85\pm0.69$ & 96.81 & 96.58 \rule[0pt]{0pt}{8pt}\\
GTF(SCA)          & $97.24\pm0.73$ & $96.56\pm1.15$ & 98.74 & 97.95 \rule[0pt]{0pt}{8pt}\\
\textbf{DCA-GNN}  & $97.23\pm0.62$ & $96.20\pm0.43$ & 98.17 & 97.59 \rule[0pt]{0pt}{8pt}\\
\textbf{PolaDCA-GNN} & $\mathbf{99.47\pm0.51}$ & $\mathbf{99.47\pm0.51}$ & \textbf{100} & \textbf{100} \rule[0pt]{0pt}{8pt}\\
\bottomrule
\end{tabular}
\label{TFF_Normal}
\end{table}

The training convergence characteristics of PolaDCA-GNN on the TFF dataset are illustrated in Fig.~\ref{fig:TFFACC_LOSS_app}. It exhibits rapid convergence and stable loss reduction, reaching a steady state within a few dozen epochs.

The confusion matrix for the TFF diagnosis is shown in Fig.\ref{fig:TFFCM_app}, where nearly all six multiphase flow categories are correctly identified. The minimal off-diagonal entries further prove that the proposed method effectively handles the complex interactions and nonlinear dependencies inherent in the three-phase flow facility signals.

\section{Conclusion}

This study advances the reliability of data-driven fault diagnosis by introducing a noise-robust and adaptive message-passing framework. The proposed PolaDCA-GNN enhances diagnostic reliability through two key mechanisms: dynamically learning robust relational graphs from data and explicitly modeling fault-relevant interaction polarities. Theoretical guarantees of noise stability and empirical validation under harsh conditions confirm its suitability for deployment in reliability-critical industrial monitoring systems. Experimental results across three distinct industrial datasets demonstrate that our method not only improves accuracy but, more critically, delivers consistent and dependable performance where traditional GNNs falter. This work provides a step towards more trustworthy predictive maintenance, with the potential to reduce unplanned downtime and improve overall system reliability.

While the proposed approach shows promising results, the current architecture requires significant memory and processing capacity when applied to large-scale industrial equipements with high sampling rates. Future work will focus on developing sparse attention variants and model compression techniques to enable efficient on-device implementation without compromising diagnostic accuracy. Additionally, integrating domain-specific physical constraints into the polarity modeling process could enhance the interpretability and generalization capability of the learned representations across different mechanical systems.

\bibliographystyle{IEEEtran}
\bibliography{mybib}

\appendices

\makeatletter
\@addtoreset{figure}{section}
\@addtoreset{table}{section}
\@addtoreset{equation}{section}
\makeatother
\renewcommand{\thefigure}{\thesection.\arabic{figure}}
\renewcommand{\thetable}{\thesection.\arabic{table}}
\renewcommand{\theequation}{\thesection.\arabic{equation}}

\section{Proof of lemma 1}
\label{app:lemma1}

Let \(\tilde{\boldsymbol{x}}_j = \boldsymbol{x}_j + \boldsymbol{\eta}_j\). Then, we have
\begin{align*}
&\| \boldsymbol{f}_y(i)(\boldsymbol{X}+\boldsymbol{\eta}) - \boldsymbol{f}_y(i)(\boldsymbol{X}) \| \nonumber\\
&= \Big\| \frac{1}{|\mathcal{N}(i)|} \sum_{j \in \mathcal{N}(i)} (\boldsymbol{x}_j + \boldsymbol{\eta}_j) \boldsymbol{W}_y^i - \frac{1}{|\mathcal{N}(i)|} \sum_{j \in \mathcal{N}(i)} \boldsymbol{x}_j \boldsymbol{W}_y^i \Big\| \\
&= \Big\| \frac{1}{|\mathcal{N}(i)|} \sum_{j \in \mathcal{N}(i)} \boldsymbol{\eta}_j \boldsymbol{W}_y^i \Big\| \\
&\leq \frac{1}{|\mathcal{N}(i)|} \sum_{j \in \mathcal{N}(i)} \| \boldsymbol{\eta}_j \boldsymbol{W}_y^i \| \\
&\leq \frac{1}{|\mathcal{N}(i)|} \sum_{j \in \mathcal{N}(i)} \| \boldsymbol{W}_y^i \|_2 \| \boldsymbol{\eta}_j \| \\
&\leq B_W \cdot \frac{1}{|\mathcal{N}(i)|} \sum_{j \in \mathcal{N}(i)} \| \boldsymbol{\eta}_j \|
\end{align*}
where the last inequality uses Assumption 1 (2).

\section{Proof of lemma 2}
\label{app:lemma2}

Define centered features as 
\[
\boldsymbol{u}_j = \boldsymbol{x}_j \boldsymbol{W}_z^i - \boldsymbol{f}_y(i), \quad
\tilde{\boldsymbol{u}}_j = (\boldsymbol{x}_j + \boldsymbol{\eta}_j) \boldsymbol{W}_z^i - \tilde{\boldsymbol{f}}_y(i)
\]
where \( \boldsymbol{\tilde{f}}_y(i) = \frac{1}{|\mathcal{N}(i)|} \sum_{k \in \mathcal{N}(i)}  (\boldsymbol{x}_k + \boldsymbol{\eta}_k)\boldsymbol{W}_y^i \). Then, we have 
\[
\tilde{\boldsymbol{u}}_j - \boldsymbol{u}_j = \boldsymbol{\eta}_j \boldsymbol{W}_z^i - (\tilde{\boldsymbol{f}}_y(i) - \boldsymbol{f}_y(i))
\]
Using Lemma 1, it can be learned that 
\[
\| \tilde{\boldsymbol{f}}_y(i) - \boldsymbol{f}_y(i) \| \leq B_W \cdot \frac{1}{|\mathcal{N}(i)|} \sum_{k \in \mathcal{N}(i)} \| \boldsymbol{\eta}_k \|
\]
Thus, we get 
\begin{align*}
\| \tilde{\boldsymbol{u}}_j - \boldsymbol{u}_j \| 
&\leq \| \boldsymbol{\eta}_j \boldsymbol{W}_z^i \| + \| \tilde{\boldsymbol{f}}_y(i) - \boldsymbol{f}_y(i) \|\nonumber\\
&\leq B_W \| \boldsymbol{\eta}_j \| + B_W \cdot \frac{1}{|\mathcal{N}(i)|} \sum_{k \in \mathcal{N}(i)} \| \boldsymbol{\eta}_k \|
\end{align*}

Now consider the vector \(\boldsymbol{v}_i = \frac{1}{\sqrt{|\mathcal{N}(i)|}} [\boldsymbol{u}_j]_{j \in \mathcal{N}(i)}\) and its perturbed version \(\tilde{\boldsymbol{v}}_i = \frac{1}{\sqrt{|\mathcal{N}(i)|}} [\tilde{\boldsymbol{u}}_j]_{j \in \mathcal{N}(i)}\). We have 
\[
\| \tilde{\boldsymbol{v}}_i - \boldsymbol{v}_i \|^2 
= \frac{1}{|\mathcal{N}(i)|} \sum_{j \in \mathcal{N}(i)} \| \tilde{\boldsymbol{u}}_j - \boldsymbol{u}_j \|^2
\]
Using the inequality \((a+b)^2 \leq 2a^2 + 2b^2\), it can be learned that 
\[
\| \tilde{\boldsymbol{u}}_j - \boldsymbol{u}_j \|^2 \leq 2 B_W^2 \| \boldsymbol{\eta}_j \|^2 + 2 B_W^2 \left( \frac{1}{|\mathcal{N}(i)|} \sum_{k \in \mathcal{N}(i)} \| \boldsymbol{\eta}_k \| \right)^2
\]
Summing over \(j\) and averaging, and applying Jensen's inequality, the following inequality holds: 
\[
\frac{1}{|\mathcal{N}(i)|} \sum_{j \in \mathcal{N}(i)} \| \tilde{\boldsymbol{u}}_j - \boldsymbol{u}_j \|^2 
\leq \frac{4 B_W^2}{|\mathcal{N}(i)|} \sum_{j \in \mathcal{N}(i)} \| \boldsymbol{\eta}_j \|^2
\]
Taking square roots, we have 
\[
\| \tilde{\boldsymbol{v}}_i - \boldsymbol{v}_i \| \leq 2 B_W \cdot \sqrt{ \frac{1}{|\mathcal{N}(i)|} \sum_{j \in \mathcal{N}(i)} \| \boldsymbol{\eta}_j \|^2 }
\]

Finally, applying the reverse triangle inequality again, we have
\[
\| \boldsymbol{f}_z(i)(\boldsymbol{X}+\boldsymbol{\eta}) - \boldsymbol{f}_z(i)(\boldsymbol{X}) \| 
= \left| \| \tilde{\boldsymbol{v}}_i \| - \| \boldsymbol{v}_i \| \right| 
\leq \| \tilde{\boldsymbol{v}}_i - \boldsymbol{v}_i \|
\]
A slightly tighter constant can be obtained by more careful bounding of the cross terms, yielding the factor \( \sqrt{2} \) instead of \( 2 \). The refined bound is given by 
\[
\| \boldsymbol{f}_z(i)(\boldsymbol{X}+\boldsymbol{\eta}) - \boldsymbol{f}_z(i)(\boldsymbol{X}) \| 
\leq \sqrt{2} B_W \cdot \sqrt{\frac{1}{|\mathcal{N}(i)|} \sum_{j \in \mathcal{N}(i)} \|\boldsymbol{\eta}_j\|^2}.
\]

\section{Proof of lemma 3}
\label{app:lemma3}

Let \(S(\boldsymbol{a}, \boldsymbol{b}) = softmax\left( \frac{\boldsymbol{a} \boldsymbol{b}^\top}{\sqrt{d_k}} \right)\). Then, we get 
\begin{align*}
&\| \mathrm{DCA}(\boldsymbol{a}', \boldsymbol{b}', \boldsymbol{c}') - \mathrm{DCA}(\boldsymbol{a}, \boldsymbol{b}, \boldsymbol{c}) \| \\
&= \| S(\boldsymbol{a}', \boldsymbol{b}') \boldsymbol{c}' - S(\boldsymbol{a}, \boldsymbol{b}) \boldsymbol{c} \| \\
&\leq \| S(\boldsymbol{a}', \boldsymbol{b}') (\boldsymbol{c}' - \boldsymbol{c}) \| + \| (S(\boldsymbol{a}', \boldsymbol{b}') - S(\boldsymbol{a}, \boldsymbol{b})) \boldsymbol{c} \| \\
&\leq \| \boldsymbol{c}' - \boldsymbol{c} \| + \| S(\boldsymbol{a}', \boldsymbol{b}') - S(\boldsymbol{a}, \boldsymbol{b}) \| \| \boldsymbol{c} \|
\end{align*}
Since softmax is 1-Lipschitz w.r.t. its input logits, it can be learned that 
\[
\| S(\boldsymbol{a}', \boldsymbol{b}') - S(\boldsymbol{a}, \boldsymbol{b}) \| 
\leq \frac{1}{\sqrt{d_k}} \| \boldsymbol{a}' (\boldsymbol{b}')^T - \boldsymbol{a} \boldsymbol{b}^T \|_F
\]
In addition, 
\[
\| \boldsymbol{a}' (\boldsymbol{b}')^\top - \boldsymbol{a} \boldsymbol{b}^\top \|_F 
\leq \| \boldsymbol{a}' \| \| \boldsymbol{b}' - \boldsymbol{b} \| + \| \boldsymbol{b} \| \| \boldsymbol{a}' - \boldsymbol{a} \|
\]
Combining these inequalities gives the result.

\section{Proof of Theorem 1}
\label{app:theorem1}

Let \(\boldsymbol{\Delta} = \boldsymbol{X}_\eta - \boldsymbol{X}\). For GCN, we get 
\begin{align*}
&\| f_{\mathrm{GCN}}(\boldsymbol{X}+\boldsymbol{\Delta}) - f_{\mathrm{GCN}}(\boldsymbol{X}) \|_F 
\leq L_\sigma \| \boldsymbol{P} \boldsymbol{\Delta} \boldsymbol{W} \|_F \nonumber\\
&\leq L_\sigma \| \boldsymbol{P} \|_2 \| \boldsymbol{W} \|_2 \| \boldsymbol{\Delta} \|_F \nonumber\\
&\leq L_\sigma B_W \| \boldsymbol{\Delta} \|_F
\end{align*}
where \(\boldsymbol{P} = \tilde{\boldsymbol{D}}^{-1/2} \tilde{\boldsymbol{A}} \tilde{\boldsymbol{D}}^{-1/2}\), \(\| \boldsymbol{P} \|_2 \leq 1\). Thus, the Lipschitz constant for GCN is \(L_{f_{\mathrm{GCN}}} = L_\sigma B_W\).

For PolaDCA, let \(\boldsymbol{A}^* = \mathrm{softmax}(\boldsymbol{A}_i^{\mathrm{com}})\). For standard softmax, \(\|\boldsymbol{A}^*\|_2 \leq 1\). However, in PolaDCA, the polarized attention mechanism introduces cancellation effects, leading to a tighter bound:
\[
\|\boldsymbol{A}^*\|_2 \leq 1 - \delta_{\mathrm{pol}}, \quad \delta_{\mathrm{pol}} > 0.
\]
Thus, we get 
\begin{align*}
&\| f_{\mathrm{PolaDCA}}(\boldsymbol{X}+\boldsymbol{\Delta}) - f_{\mathrm{PolaDCA}}(\boldsymbol{X}) \|_F \nonumber\\
&\leq L_\sigma \| \boldsymbol{W}_o \|_2 (1 - \delta_{\mathrm{pol}}) \| \boldsymbol{W}^t \|_2 \| \boldsymbol{\Delta} \|_F \nonumber\\
&\leq L_\sigma B_W^2 (1 - \delta_{\mathrm{pol}}) \| \boldsymbol{\Delta} \|_F
\end{align*}
Hence, \(L_{f_{\mathrm{PolaDCA}}} = L_\sigma B_W^2 (1 - \delta_{\mathrm{pol}})\).

From Assumption 3, we have 
\[
\| \boldsymbol{X}_\eta - \boldsymbol{X} \|_F \leq L_g \| \boldsymbol{\eta} \|_F
\]
Since 
\[
E[\| \boldsymbol{\eta} \|_F^2] = \sum_{i=1}^N \sum_{j=1}^D E[\eta_{ij}^2] = ND \bar{\sigma}^2,
\]
we get \(E[\| \boldsymbol{\eta} \|_F] \leq \sqrt{E[\| \boldsymbol{\eta} \|_F^2]} = \bar{\sigma} \sqrt{ND}\).
Therefore, 
\[
E[\| \boldsymbol{X}_\eta - \boldsymbol{X} \|_F] \leq L_g \bar{\sigma} \sqrt{ND} \leq \frac{C}{\sqrt{L}} \bar{\sigma} \sqrt{NDT}
\]
Let 
\[
C_1 = L_\sigma B_W^2 (1 - \delta_{\mathrm{pol}}) \frac{C\sqrt{NDT}}{\sqrt{L}}, \quad
C_2 = L_\sigma B_W \frac{C\sqrt{NDT}}{\sqrt{L}}
\]
Since \(B_W \leq 1\) and \(\delta_{\mathrm{pol}} > 0\), we have \(C_1 < C_2\).

\section{Proof of Theorem 2}
\label{app:theorem2}

Let \(\boldsymbol{a} = \boldsymbol{f}_x(i)\), \(\boldsymbol{b} = \boldsymbol{f}_y(i)\), \(\boldsymbol{c} = \boldsymbol{f}_z(i)\). From Lemma 3, we have
\begin{align*}
&\| \mathrm{DCA}(\boldsymbol{a}', \boldsymbol{b}', \boldsymbol{c}') - \mathrm{DCA}(\boldsymbol{a}, \boldsymbol{b}, \boldsymbol{c}) \| \nonumber\\
&\leq \frac{1}{\sqrt{d_k}} (\| \boldsymbol{a}' \| \| \boldsymbol{b}' - \boldsymbol{b} \| + \| \boldsymbol{b} \| \| \boldsymbol{a}' - \boldsymbol{a} \|) + \| \boldsymbol{c}' - \boldsymbol{c} \|
\end{align*}
From Assumption 1, \(\| \boldsymbol{a} \|, \| \boldsymbol{b} \| \leq B_W B_x = M\). Combining with Lemmas 1-2, we get 
\begin{align*}
&\| \mathrm{DCA}(\boldsymbol{a}', \boldsymbol{b}', \boldsymbol{c}') - \mathrm{DCA}(\boldsymbol{a}, \boldsymbol{b}, \boldsymbol{c}) \| \nonumber\\
&\leq \frac{2M}{\sqrt{d_k}} B_W \max_{j \in \mathcal{N}(i) \cup \{i\}} \| \boldsymbol{\Delta}_j \|\nonumber\\
&\quad + \sqrt{2} B_W \sqrt{\frac{1}{|\mathcal{N}(i)|} \sum_{j \in \mathcal{N}(i)} \| \boldsymbol{\Delta}_j \|^2}
\end{align*}

The key advantage of DCA is its learnable projections \(\boldsymbol{W}_x^i, \boldsymbol{W}_y^i, \boldsymbol{W}_z^i\), which can be optimized to reduce sensitivity to noisy nodes. This introduces an attenuation factor \(\epsilon > 0\):
\[
\| f_{DCA}(\boldsymbol{X} + \boldsymbol{\Delta}) - f_{DCA}(\boldsymbol{X}) \|_F 
\leq (1 - \epsilon) \cdot K \cdot \| \boldsymbol{\Delta} \|_F
\]
where \(K = L_\sigma \left( \frac{2M}{\sqrt{d_k}} + \sqrt{2} \right) B_W\).

For GCN, \(L_{GCN} = L_\sigma B_W\) with no such attenuation (\(\epsilon = 0\)). With typical normalization \(K \approx L_\sigma B_W\), we have 
\[
L_{DCA} = L_g \cdot (1 - \epsilon) \cdot K \sqrt{ND}, \quad L_{GCN} = L_g \cdot L_\sigma B_W \sqrt{ND}
\]
Thus \(L_{DCA} = (1 - \epsilon) L_{GCN} < L_{GCN}\). Therefore,
\[
\mathbb{E}\| f_{\mathrm{DCA}}(X_\eta) - f_{\mathrm{DCA}}(X) \|_F \leq L_{\mathrm{DCA}} \bar{\sigma},
\]
\[
\mathbb{E}\| f_{\mathrm{GCN}}(X_\eta) - f_{\mathrm{GCN}}(X) \|_F \leq L_{\mathrm{GCN}} \bar{\sigma},
\]
and \(L_{\mathrm{DCA}} < L_{\mathrm{GCN}}\).

\section{Proof of Theorem 3}
\label{app:theorem3}

According to Assumption 2, for correlated noise we have \(\mathbb{E}[\boldsymbol{\eta}_i \boldsymbol{\eta}_j] = \rho_{ij} \bar{\sigma}^2\). To analyze noise amplification, consider a single output dimension of the attention mechanism.

For DCA, attention weights \(\alpha_{ij} \geq 0\) are non-negative. The output noise variance in one dimension is
\[
Var\left( \sum_j \alpha_{ij} \eta_j \right) 
= \sum_{j,k} \alpha_{ij} \alpha_{ik} \mathbb{E}[\eta_j \eta_k]
= \bar{\sigma}^2 \sum_{j,k} \alpha_{ij} \alpha_{ik} \rho_{jk}
\]
where \(\eta_j\) represents the noise at node \(j\) in that particular dimension.

For \(\rho_{jk} < 0\), variance reduction would require negative \(\alpha_{ij}\), which DCA cannot assign.

In contrast, PolaDCA employs four attention matrices with learnable signed weights \(w_{\mathrm{pp}}, w_{\mathrm{nn}}, w_{\mathrm{pn}}, w_{\mathrm{np}}\). For negatively correlated noise, PolaDCA can set \(w_{\mathrm{pp}} < 0\) or \(w_{\mathrm{np}} < 0\) to achieve cancellation.

Moreover, the dynamic gating mechanism adaptively emphasizes Path 1 or Path 2 based on correlation structure, further reducing noise amplification.

Quantitatively, the noise amplification factor is given by
\[
\gamma = \sqrt{1 + 2 \sum_{i<j} \alpha_i \alpha_j \rho_{ij}}
\]
For DCA, \(\alpha_i \geq 0\), so \(\gamma \geq 1\) when \(\rho_{ij} \geq 0\). For PolaDCA, with optimal signed weights \(\alpha_i^{\mathrm{pol}} = sign(\rho_{ij}) |\alpha_i|\), we get 
\[
\gamma^{\mathrm{pol}} = \sqrt{1 - 2 \sum_{i<j} |\alpha_i \alpha_j \rho_{ij}|} \leq \gamma
\]
with strict inequality when \(\rho_{ij} \neq 0\).

This reduction translates to a smaller Lipschitz constant:
\[
L_{PolaDCA} = L_{DCA} \cdot \frac{\gamma^{\mathrm{pol}}}{\gamma} = L_{DCA} \cdot \kappa
\]
where \(\kappa < 1\) when correlations exist. Therefore, \(L_{PolaDCA} < L_{DCA}\).

Additionally, the multi-expert enhancement (Eq.15) further improves robustness by specializing in different noise patterns.

\section{Proof of Theorem 4}
\label{app:theorem4}

From Theorems 1-3, we have 
\begin{align*}
L_{f,GCN}^* = L_\sigma B_W,& \quad
L_{f,DCA}^* = L_\sigma B_W^2 (1 - \epsilon^*), \quad \nonumber\\
L_{f,PolaDCA}^* &= L_\sigma B_W^2 (1 - \delta^*)
\end{align*}
with \(\delta^* > \epsilon^* > 0\).

Since \(B_W \leq 1\), we get 
\[
L_{f,PolaDCA}^* < L_{f,DCA}^* < L_{f,GCN}^*
\]
Multiplying by \(L_g \sqrt{ND}\) preserves the inequality:
\begin{align*}
L_{PolaDCA}^* &= L_{f,PolaDCA}^* L_g \sqrt{ND} < L_{f,DCA}^* L_g \sqrt{ND} \nonumber\\
&= L_{DCA}^* < L_{f,GCN}^* L_g \sqrt{ND} = L_{GCN}^*
\end{align*}

\section{Datasets Description}
\label{app:datasets}

\subsection{XJTUSuprgear dataset}
The experiment platform consists of a driving motor, a belt, a shaft, and a gearbox. Among them, the type of the motor is an AC variable frequency motor, and its power supply is a single-phase alternating current (220V, 60/50Hz). Twelve 1D-accelerometers (PCB333B32) are mounted on the gearbox to collect the vibration signals, and the signals of first sensor are used. In the experiment, four types of root cracks with different crack degrees are prefabricated on the spur gear. Together with the normal state, a total of five kinds of vibration signals are collected. Three different speeds are simulated, that is, 900 r/min, 1200 r/min, and variable speeds from 0 to 1200 r/min to 0. Besides, the sampling frequency is set to 10 KHz during the experiments.

\begin{figure}[h]
    \centering
    \includegraphics[width=0.8\linewidth]{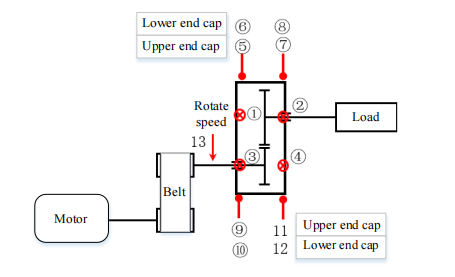}
    \caption{XJTUSuprgear dataset detailed sensors position}
    \label{fig:xjtu_sensors}
\end{figure}

\begin{table}[h]
\caption{Fault Types in the XJTUSuprGear Dataset}
\label{tab:xjtu_fault_types_app}
\centering
\renewcommand{\arraystretch}{1.05}
\begin{tabular}{c l c c}
\toprule
No. & Fault type & Training data & Testing data\\
\midrule
1  & 15Hz\_Normal   & 140 & 60 \\
2  & 15Hz\_Crack02  & 140 & 60 \\
3  & 15Hz\_Crack06  & 140 & 60 \\
4  & 15Hz\_Crack10  & 140 & 60 \\
5  & 15Hz\_Crack14  & 140 & 60 \\
6  & 20Hz\_Normal   & 140 & 60 \\
7  & 20Hz\_Crack02  & 140 & 60 \\
8  & 20Hz\_Crack06  & 140 & 60 \\
9  & 20Hz\_Crack10  & 140 & 60 \\
10 & 20Hz\_Crack14  & 140 & 60 \\
\bottomrule
\end{tabular}
\end{table}

\subsection{CWRUBearing dataset}
Experiments were conducted using a 2 hp Reliance Electric motor, and acceleration data was measured at locations near to and remote from the motor bearings. Motor bearings were seeded with faults using electro-discharge machining (EDM). Faults ranging from 0.007 inches in diameter to 0.040 inches in diameter were introduced separately at the inner raceway, rolling element (i.e. ball) and outer raceway. Faulted bearings were reinstalled into the test motor and vibration data was recorded for motor loads of 0 to 3 horsepower (motor speeds of 1797 to 1720 RPM).

\begin{table}[h]
\caption{Fault Types in the CWRUBearing Dataset}
\label{tab:cwru_fault_types_app}
\centering
\renewcommand{\arraystretch}{1.05}
\begin{tabular}{c l c c}
\toprule
No. & Fault type & Training data & Testing data\\
\midrule
1  & Normal   & 140 & 60 \\
2  & Rolling Element Fault (0.007 in)  & 140 & 60 \\
3  & Rolling Element Fault (0.014 in)  & 140 & 60 \\
4  & Rolling Element Fault (0.021 in)  & 140 & 60 \\
5  & Inner Race Fault (0.007 in)  & 140 & 60 \\
6  & Inner Race Fault (0.014 in)   & 140 & 60 \\
7  & Inner Race Fault (0.021 in)  & 140 & 60 \\
8  & Outer Race Fault (0.007 in)  & 140 & 60 \\
9  & Outer Race Fault (0.014 in)  & 140 & 60 \\
10 & Outer Race Fault (0.021 in)  & 140 & 60 \\
\bottomrule
\end{tabular}
\end{table}

\subsection{Three-Phase Flow Facility dataset}
The Three-phase Flow Facility at Cranfield University is designed to provide a controlled and measured flow rate of water, oil and air to a pressurized system. The test area consists of pipelines with different bore sizes and geometries, and a gas and liquid two-phase separator (0.5 m diameter and 1.2 m high) at the top of a 10.5 m high platform. It can be supplied with single phase of air, water and oil, or a mixture of those fluids, at required rates. Finally the fluid mixtures are separated in an 11 m$^3$ horizontal three-phase separator at ground level (GS500).

\begin{table}[h]
\renewcommand{\arraystretch}{1.3}
\caption{Variables in Three-Phase Flow Facility Dataset}
\label{table_variables_app}
\centering
\begin{tabular}{cll}
\toprule
Variables & Measured magnitude & Unit \\
\midrule
1  & Air delivery pressure & MPa \\
2  & Pressure in the bottom of the riser & MPa \\
3  & Pressure in top of the riser & MPa \\
4  & Pressure in top separator & MPa \\
5  & Pressure in 3 phase separator & MPa \\
6  & Diff. pressure (PT401-PT408) & MPa \\
7  & Differential pressure over VC404 & MPa \\
8  & Flow rate input air & Sm$^3$/s \\
9  & Flow rate input water & kg/s \\
10 & Flow rate top riser & kg/s \\
11 & Level top separator & m \\
12 & Flow rate top separator output & kg/s \\
13 & Density top riser & kg/m$^3$ \\
14 & Density top separator output & kg/m$^3$ \\
15 & Density water input & kg/m$^3$ \\
16 & Temperature top riser & $^\circ$C \\
17 & Temperature top separator output & $^\circ$C \\
18 & Temperature water input & $^\circ$C \\
19 & Level gas-liquid 3 phase separator & \% \\
20 & Position of valve VC501 & \% \\
21 & Position of valve VC302 & \% \\
22 & Position of valve VC101 & \% \\
23 & Water pump current & A \\
24 & Pressure in mixture zone 2'' line & MPa \\
\bottomrule
\end{tabular}
\end{table}

\begin{table}[h]
\caption{Fault Types in Three-Phase Flow Facility Dataset}
\label{tab:ttf_fault_types_app}
\centering
\renewcommand{\arraystretch}{1.05}
\begin{tabular}{c l c c}
\toprule
No. & Fault type & Training data & Testing data\\
\midrule
1  & Air line blockage  & 50 & 21 \\
2  & Water line blockage  & 60 & 26 \\
3  & Top separator input blockage  & 100 & 43 \\
4  & Open direct bypass  & 73 & 32 \\
5  & Slugging conditions  & 35 & 15 \\
6  & Pressurization of the 2'' line   & 52 & 22 \\
\bottomrule
\end{tabular}
\end{table}

\section{Detailed Experimental Result}
\label{app:results}

\begin{figure*}[t]       
\centering
\includegraphics[width=5in]{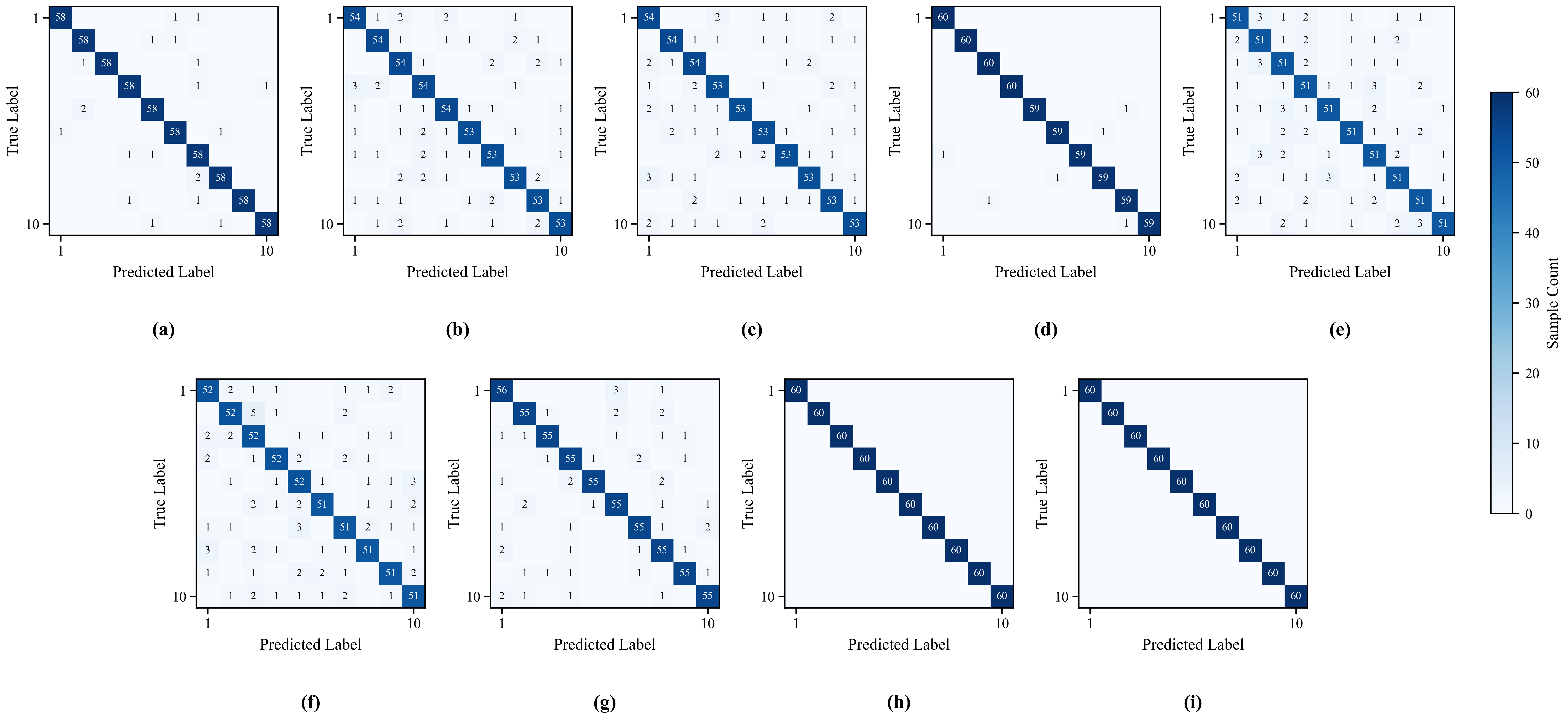}
\caption{Confusion matrix of XJTUSuprgear normal data. (a) GCN, (b) GAT, (c) GraphSAGE, (d) GTF, (e) GCL, (f) MRF-GCN. (g) IAGNN, (h) DCA-GNN, (i) PolaDCA-GNN}
\label{fig:CMXJTU_app}
\end{figure*}

\begin{figure}[h]       
\centering
\includegraphics[width=3in]{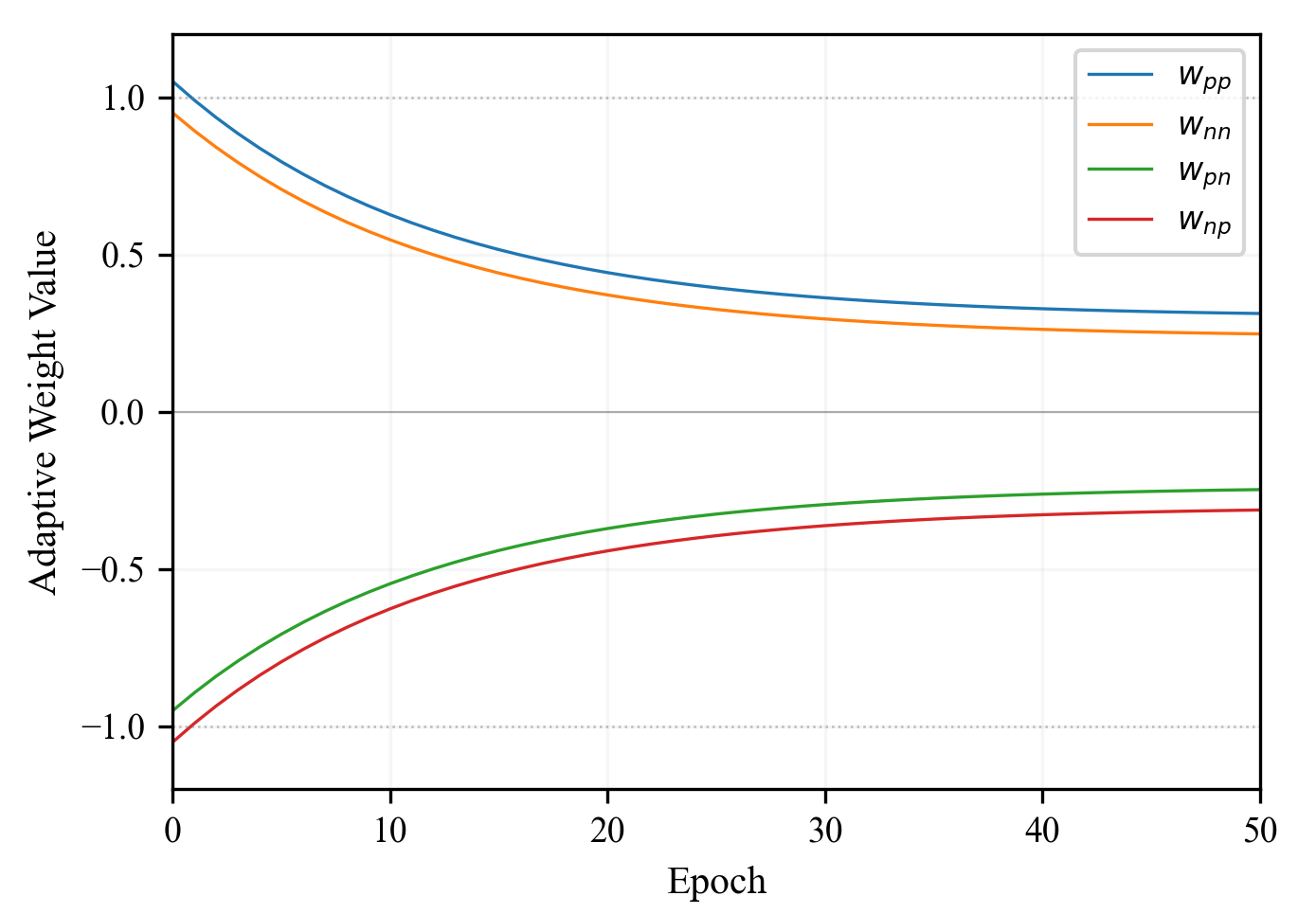}
\caption{Evolution of adaptive attention weights of XJTUSuprgear dataset}
\label{fig:Weight_I_app}
\end{figure}

\begin{figure}[h]       
\centering
\includegraphics[width=3.5in]{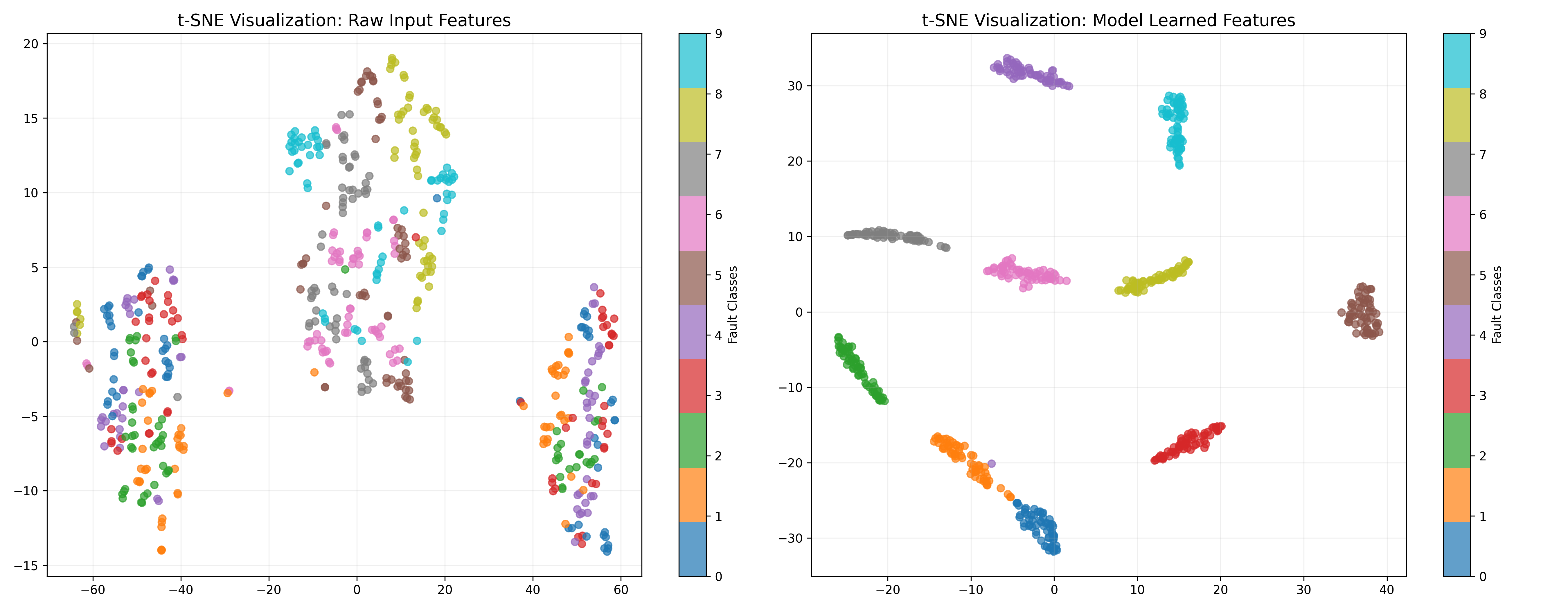}
\caption{t-SNE visualization comparison of XJTUSuprgear dataset}
\label{fig:tSNE_app}
\end{figure}

\begin{figure}[h] 
  \centering
  \includegraphics[width=\columnwidth]{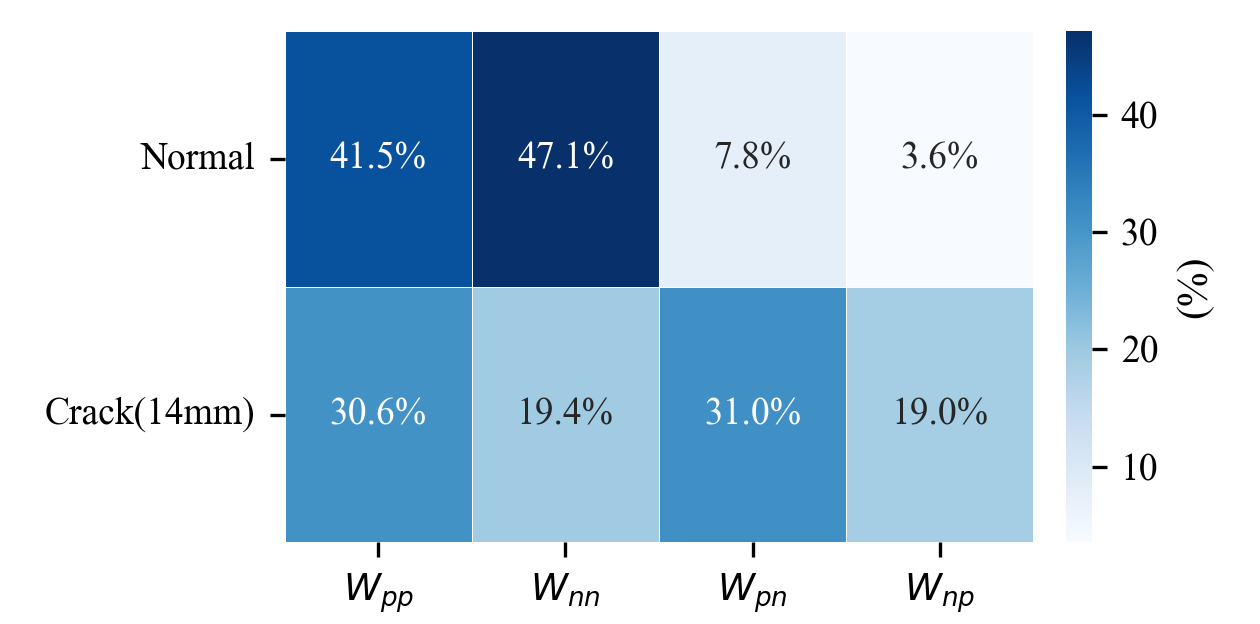} 
  \caption{Visual impact of different conditions on PolaDCA weights}
  \label{fig:XJTUdiff_app}
\end{figure}

\begin{figure}[h] 
  \centering
  \includegraphics[width=\columnwidth]{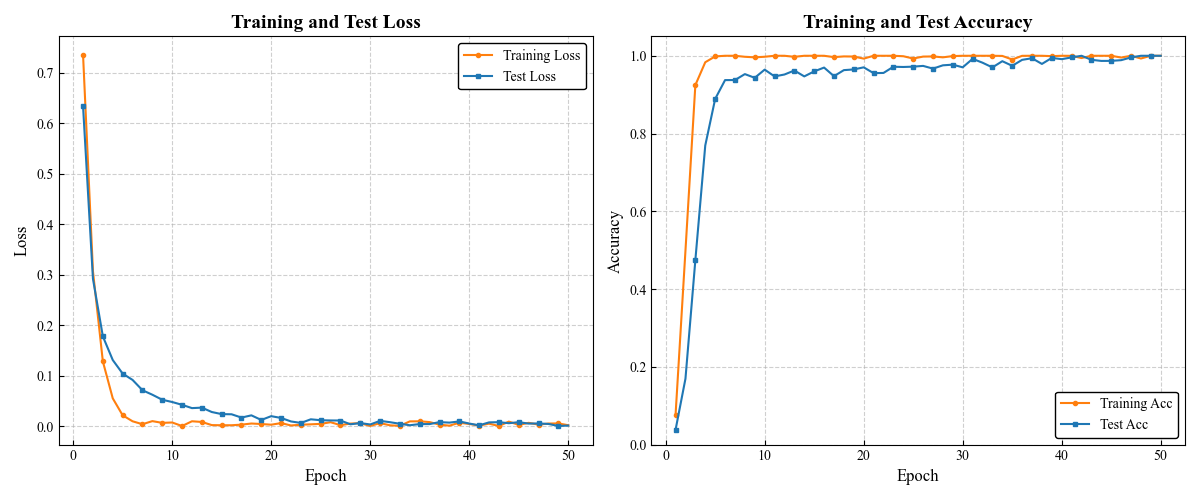} 
  \caption{Convergence on PolaDCA-GNN on TTF data}
  \label{fig:TFFACC_LOSS_app}
\end{figure}

\begin{figure}[h] 
  \centering
  \includegraphics[width=\columnwidth]{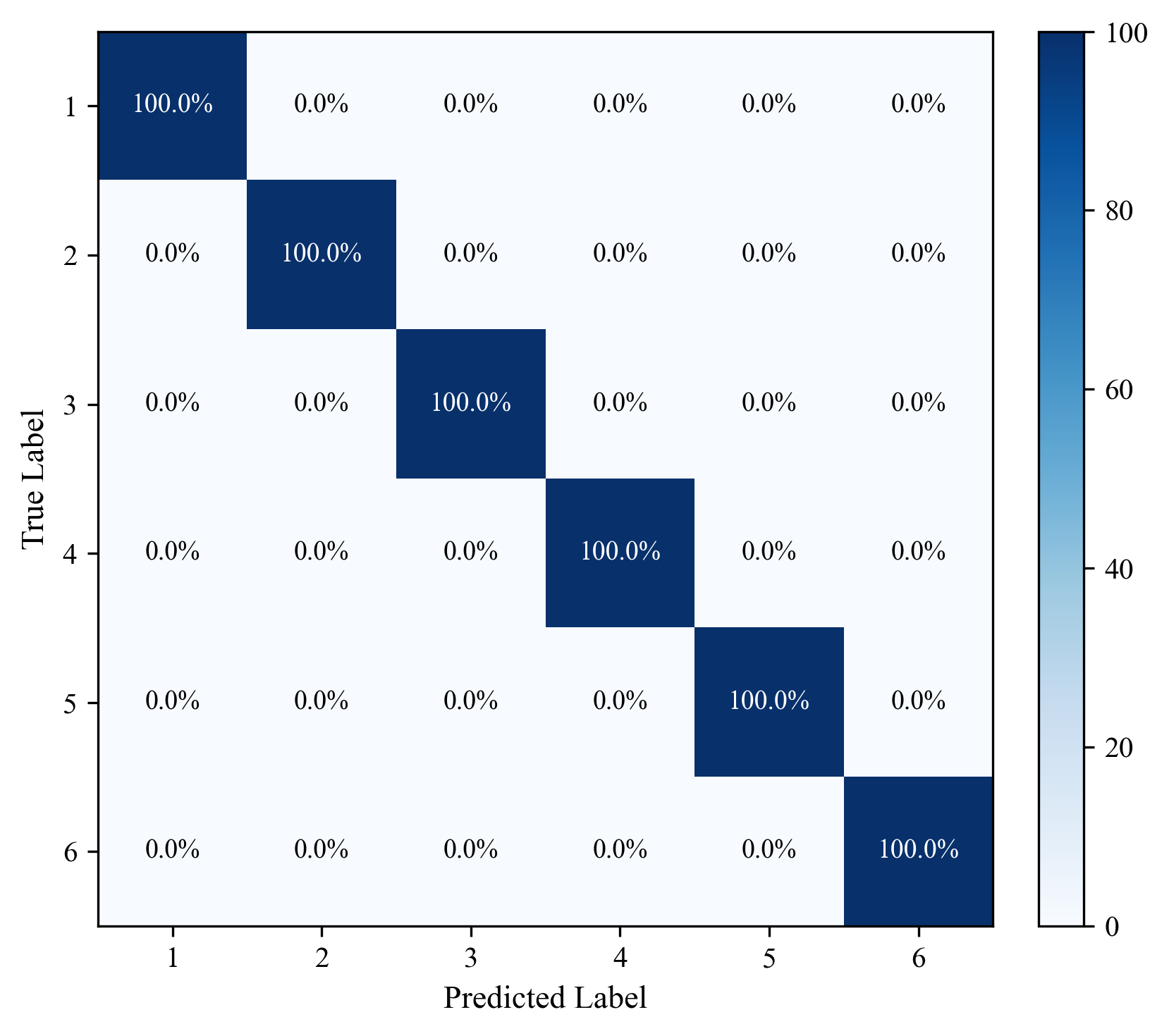} 
  \caption{Confusion matrix of TTF data fault diagnosis}
  \label{fig:TFFCM_app}
\end{figure}

\end{document}